\documentclass{article}
\usepackage{arxiv}

\usepackage[utf8]{inputenc} 
\usepackage{natbib}

\usepackage[T1]{fontenc}    
\usepackage{hyperref}       
\usepackage{booktabs}       

\usepackage{graphicx}
\usepackage{caption}
\usepackage{subcaption}
\usepackage{array}
\usepackage{amsmath}
\usepackage{amssymb}
\usepackage{amsfonts}
\usepackage{bm}
\usepackage{xcolor}
\usepackage{multirow,array}
\usepackage{comment}
\usepackage{url} 
\usepackage{tikz}
\usepackage{pgfplots}
\usepackage[normalem]{ulem}
\usetikzlibrary{matrix,positioning}
\usepackage{makecell}
\usepackage[flushleft]{threeparttable}
\usepackage{multicol}

\DeclareSymbolFont{bbold}{U}{bbold}{m}{n}
\DeclareSymbolFontAlphabet{\mathbbold}{bbold}

\definecolor{luisas_color}{HTML}{5c4ae4}
\definecolor{ann_color}{HTML}{42e6f5}

\definecolor{fredrik_color}{HTML}{234567}

\newtheorem{Def}{Definition}

\title{A practical guide to multi-objective reinforcement learning and planning}

\author{
  Conor F. Hayes\thanks{These authors contributed equally to this work.} \\
  National University of Ireland Galway\\
Galway, Ireland\\
  \texttt{{c.hayes13@nuigalway.ie}} \\
  \And
  Roxana R\u{a}dulescu\footnotemark[1]\\
 Vrije Universiteit Brussel \\
Brussels, Belgium\\
  \texttt{{roxana.radulescu@vub.be}} \\
  \And
  Eugenio Bargiacchi \\
Vrije Universiteit Brussel \\
Brussels, Belgium\\
  \And 
  Johan K\"allstr\"om \\
  Linköping University\\
    Linköping, Sweden\\
  \And 
  Matthew Macfarlane \\
  University of Amsterdam\\
              the Netherlands\\
  \And  
  Mathieu Reymond \\
  Vrije Universiteit Brussel \\
Brussels, Belgium\\
  \And 
  Timothy Verstraeten \\
  Vrije Universiteit Brussel \\
Brussels, Belgium\\
\And 
Luisa M.\ Zintgraf \\
University of Oxford\\
              Oxford, United Kingdom\\
\And 
Richard Dazeley \\ 
 Deakin University\\
              Geelong, Australia\\
\And 
Fredrik Heintz \\ 
 Linköping University\\
    Linköping, Sweden\\
\And  
Enda Howley \\
National University of Ireland Galway\\
              Galway, Ireland\\
\And 
Athirai A. Irissappane \\
University of Washington\\
Tacoma, USA\\
\And 
Patrick Mannion \\
National University of Ireland Galway\\
              Galway, Ireland\\
\And 
Ann Now\'{e}\\
Vrije Universiteit Brussel \\
Brussels, Belgium\\
\And 
Gabriel Ramos \\
   Universidade do Vale do Rio dos Sinos\\
              S\~ao Leopoldo, RS, Brazil\\
\And 
Marcello Restelli  \\
Politecnico di Milano\\
              Milan, Italy\\
\And 
Peter Vamplew \\
Federation University Australia\\
              Ballarat, Australia\\
\And
Diederik M.\ Roijers\\
Vrije Universiteit Brussel, Belgium  \& \\
              HU University of Applied Sciences Utrecht,  
              the Netherlands\\
}

\begin{document}
\maketitle

\begin{abstract}
Real-world decision-making tasks are generally complex, requiring trade-offs between multiple, often conflicting, objectives. Despite this, the 
majority of research in reinforcement learning and decision-theoretic planning either assumes only a single objective, or that multiple objectives can be adequately handled via a simple linear combination. Such approaches may oversimplify the underlying problem and hence produce suboptimal results. This paper serves as a guide to the application of multi-objective methods to difficult problems, and is aimed at researchers who are already familiar with single-objective reinforcement learning and planning methods who wish to adopt a multi-objective perspective on their research, as well as practitioners who encounter multi-objective decision problems in practice. It identifies the factors that may influence the nature of the desired solution, and illustrates by example how these influence the design of multi-objective decision-making systems for complex problems. 
\end{abstract}


\section{Introduction}
\label{sec:introduction}
In most real-world decision problems, we care about more than one aspect. For example, if we have a water reservoir with a hydro-electric power plant we may care about maximising energy production, while minimising irrigation deficits as well as minimising the risk of flooding \citep{reddy2006optimal,pianosi2013tree,castelletti2013multiobjective}. In medical treatment, we may want to maximise the effectiveness of the treatment, while minimising a variety of side effects \citep{jalalimanesh2017multi,laber2014set,lizotte2010efficient}.
In other words, most real-world decision problems are inherently multi-objective. 

While most decision problems actually have multiple objectives, most algorithms dealing with agents that need to interact with decision problems focus on optimising a single objective \citep{sutton2018reinforcement}. To nonetheless deal with the multiple objectives of the real world, a common approach to creating decision-theoretic agents is to combine all the important aspects together into a single, scalar, additive reward function. This typically involves an iterative process of assigning numerical rewards or penalties  to events that can occur in the environment. For example, in the water reservoir setting, we may put a large penalty on a flood occurring, a positive reward on the power output for each time step, and a negative reward for each time step in which the irrigation demand is not met. Then, the single-objective planning or learning agent is turned on, the resulting policy observed, and then the reward function is re-engineered if the behaviour is not satisfactory. This iterative process is then repeated until the behaviour is acceptable to the designer. We argue that this workflow is problematic for several reasons, which we will discuss in detail one by one: (a)~it is a semi-blind manual process,
 (b)~it prevents people who should take the decisions from making well-informed trade-offs, putting an undue burden on engineers to understand the decision-problem at hand, (c)~it damages the explainability of the decision-making process, and (d)~it cannot handle all types of preferences that users and human decision makers might actually have. Finally, (e)~preferences between objectives may change over time, and a single-objective agent will have to be retrained or updated when this happens. 
 
Firstly (a), if we engineer a scalar reward function through an iterative process until we reach acceptable behaviour, we try out multiple reward functions, each of which is a \emph{scalarisation} of the actual objectives. However, we do not systematically inspect all possible reward functions. In other words, we may meet our minimal threshold for acceptable behaviours, but we only observe a subset of all possible scalarisations. Therefore, although an acceptable solution may be found, it can be arbitrarily far away from optimal utility -- the use we would have received if we could have systematically examined all possible solutions. This automatically brings us to the second point (b). As the reward function is something that needs to be engineered beforehand, we are only guessing as to the effects this might have on the policy. For example, when trying to train an agent in a power production system, we may wish to double the average power output. 
However, even if the objectives are linearly weighted in the reward function, it is not as simple as just doubling the reward associated with the power output aspect of performance, as the relationship between the reward weights and the actual objective outcomes may well be non-linear \citep{van2014novel}. If, on the other hand, we would be able to inspect all \emph{possibly optimal} policies -- and their values offering different trade-offs between the objectives -- we could have decided in a well-informed manner on the outcomes, rather than making an educated guess at the scalarisation a priori. This educated guessing is also putting decision power where it does not belong: with the engineers. When an engineer creates a scalar reward function, they are simultaneously making assumptions about the preferences of the actual decision makers (e.g., a government in case of the water reservoir) and making guesses about the behavioural changes resulting from changes to the scalar reward function. This is not a responsibility that can be left to AI engineers -- at least not in decision problems that are of significant importance. 

Another issue with scalar reward functions is the lack of (a posteriori) explainability (c). If we ask ``why did the robot collide with and destroy the vase?'', we could try to input an alternative decision, such as swerving away from the vase. An agent with a single all-encompassing objective that has learnt a scalar value function will then, for example, tell us there was a $3.451$ reduction in value for this other policy, which provides little insight.

If instead, the agent could have told us that in the objective of damage to property the probability of damaging the vase would have dropped to practically $0$, but the probability of running into the family dog increased by $0.5\%$ (a different objective), this would give us insight into what went wrong. We may also disagree for different reasons: we may think that the agent has overestimated the risk of colliding with the dog, which would be an error in the value-estimate for that objective. We might also think that a $0.5\%$ increase in the likelihood of bumping into to the dog would be so small that it would have been acceptable -- especially if the robot bumping into the dog would probably have been an inconvenience for the dog, but not an actual danger to it -- if the robot could have definitely avoided destroying the vase. This would have been an error in the utility function we assign to different outcomes. In other words, not taking an explicitly multi-objective approach can rob us of essential information that we might need to evaluate or understand our agents. 

Furthermore (d), not all human preferences can be handled by scalar additive reward functions \citep{roijers2013survey}. When a user's preferences ought to be modelled with a non-linear rather than a linear utility function, a priori scalarisation becomes mathematically impossible within many reinforcement learning frameworks, as scalarisation would break the additivity of the reward function. For some domains, this might still be acceptable, as the resulting loss of optimality may not have a major impact. However, in important domains where ethical or moral issues become apparent, single-objective approaches require explicitly combining these factors together with other objectives (such as economic outcomes) in a way that may be unacceptable to many people \citep{wallach2008moral}. Similarly, designing single-objective rewards may be difficult or impossible for scenarios where we wish to ensure fair or equitable outcomes for multiple participants \citep{vamplew2018,siddiquelearning}.

Finally (e), humans are known to change their minds from time to time. Therefore, preferences between trade-offs in the different objectives may well change over time. An explicitly multi-objective system can train agents to be able to handle such preference changes, thereby preempting the need to discover a new policy whenever such changes occur. This increases the applicability of multi-objective decision-making agents, as agents do not need to be taken out of operation to be updated and they can simply switch policy to match the new user preferences. We note that this type of change is different from the issue of non-stationary dynamics of the problem which can occur in both single-objective and multi-objective problems; here the
multi-objective Markov decision process (Section \ref{sec:bg}) itself is stationary, but the external preferences have changed. 

An insight into the difference between single-objective and multi-objective approaches to an application can be gained by comparing two different studies applying RL to wet clutch engagement \citep{van2012improving,brys2013behaviour}. The task is to control the piston in a wet clutch so as to produce a fast and smooth engagement, by minimising both the time to engagement and the torque loss. The initial study uses a scalar reward with discounting which implicitly captures both aspects of the desired behaviour and achieves acceptable results \citep{van2012improving}. However, the subsequent study examines the policies arising from several different utility functions and parameterisations of those functions and demonstrates that some of these are superior to those reported in the original work \citep{brys2013behaviour}.

By now, we hope we have convinced you, the reader, that taking an explicitly multi-objective approach to planning and learning may be essential to deploying AI in decision problems. To provide further motivation, as well as showcase some difficulties that can arise when modelling problems with multiple objectives, we will provide examples of such multi-objective decision problems in Section~\ref{sec:motivating_examples}. We then proceed with formalising multi-objective problems (Section~\ref{sec:bg}) and recommend an approach to systematically deal with multi-objective decision problems that puts the user's utility front-and-centre throughout the entire process (Section~\ref{sec:utility_based}). In Section \ref{sec:design_factors} we outline which factors should be taken into account in the process from identifying a multi-objective decision problem to deploying a policy for it in practice. We describe the effects of these factors on both this process and on the solution concepts. We then proceed to describe the relationships between multi-objective decision problems and other known decision problems (Section \ref{sec:relationships}), and briefly survey both algorithmic approaches (Section \ref{sec:survey_algorithms}) and the metrics for evaluating the solutions produced by these algorithms (Section \ref{sec:evaluating_performance}).
To help researchers get started in the field, we include a worked-out example of a multi-objective decision problem, a water management problem with multiple objectives, in Section \ref{sec:worked_examples}, furthermore, we added a Jupyter notebook \citep{jupyter} with these worked-out examples as supplementary material. Finally, we conclude the article and discuss open research challenges in Section \ref{sec:challenges_open}. 

Our purpose with this article is to provide an introduction to multi-objective decision making and guide the reader through getting started with modelling and solving such decision problems. 
This article differs from existing surveys in the literature that aim to provide a comprehensive overview of methods and theory, in that it is designed to be a guide for practitioners and researchers, highlighting the issues that need to be considered and addressed when applying multi-objective agents to practical problems. As a follow-on reading, we recommend the more technical survey provided by \citet{roijers2013survey}.

\section{Motivating examples of modelling complex problems with multi-objective approaches}
\label{sec:motivating_examples}
This section presents examples of complex decision-making situations where multi-objective approaches play a role. These examples motivate some of the aspects discussed in later sections.

\subsection{Planning a journey}
\label{sec:journey}

Consider you need to travel from your house to a given destination. Deciding on the modes of transportation typically involves a number of objectives, such as minimising travel time and cost whereas maximising comfort and reliability \citep{Ortuzar&Willumsen2011,Ramos+2020aamas,Mannion+2016chap}. For instance, car trips may be faster and more comfortable than subway ones, at the cost of being more expensive and less reliable (at least in cities that easily get congested due to e.g. an accident). Moreover, given the competitive nature of traffic, your objectives are usually affected by other users, which increases the uncertainties associated with your decision. In spite of such uncertainties, if you can express your preferences over these different objectives as a linear combination, then you can make your decision using conventional optimisation approaches. However, if (as is often the case) you cannot articulate your preferences explicitly in a single formula, or you actually can, but this formula is non-linear, 
 you have a genuine multi-objective problem, which requires a multi-objective approach (see details in Section~\ref{sec:seresr}). 

In order to select the best multi-objective approach, different factors come into play. If you execute this journey every day, you might be interested in balancing your objectives on average over a longer period. However, your intention might also be to balance the objectives during each of the single journeys, which would require a different approach. Both views would result in one policy that tells you how to plan your journey. Nonetheless, at some occasions, you might want to balance the objectives differently because you have an important meeting or you have someone accompanying you on your journey. If you want to be prepared for this, you can apply a method that provides you with a variety of policies, each of which is optimising a different combination of the objectives involved. In this situation, you could easily adjust each single trip based on your current needs. In contrast, conventional optimisation approaches would need to recompute the policy from scratch in order to handle such changes. 

\subsection{Water management}
\label{sec:water}

Water reservoir operations need to handle multiple competing objectives related to significant socio-economic impacts \citep{castelletti2008water}. By regulating a system of dam gates placed at the outlet of a lake you can modulate the water release and the level of the lake. On the one hand, you will need to supply water to downstream users to meet their agricultural needs. To achieve this, you need to store water during the winter and spring in order to release it during the irrigation season. On the other hand, stakeholders on the shores are interested in keeping the lake level within a certain range to avoid floods and support recreational activities or environmental services. Increasing the lake storage to avoid irrigation deficits means increasing the risk of flooding and therefore some compromise needs to be established. The regulation problem is complicated by the presence of other objectives that interact with the two above: hydropower production, flood mitigation for downstream users, lake navigability, and many others \citep{reddy2006optimal,pianosi2013tree,castelletti2013multiobjective}.
A multi-objective analysis is a fundamental tool for the human operator and for the representatives of the various stakeholders to properly evaluate the possible trade-offs among the several conflicting objectives and to support their decisions.

\subsection{Military purchasing}
\label{sec:military}

The manufacture and purchasing of military equipment requires long term dynamic planning  \citep{nguyena2017hybrid}. Each type of equipment takes a varying degree of time to manufacture. For example, a truck may only need a week while a submarine may need more than ten years. Furthermore, the time and cost in setting up a manufacturing pipeline will require items to be produced in larger numbers. Governments need to make decisions now based on a prediction of the types of environments and operations they expect to deploy forces to in the future. These environments typically require unique combinations of equipment to maximise their ability to achieve the outcomes required. Determining this optimum combination of equipment required for operations ten to fifteen years into the future is a multi-objective planning problem -- weighing-up various factors such as the cost, effectiveness, versatility, and protection provided to personnel. In practice, this becomes a problem with many objectives, when considering details such as the selected features of each piece of equipment. For instance, \citet{beliakov2019multicriteria} discusses some seventeen objectives (related to survivability, lethality, mobility, and knowledge) to be considered when simply purchasing a single tank. 

Furthermore, in the real world, any initial decision made must also be constantly altered over subsequent years. These alterations may be instigated by a change in: government; national priorities; international dynamics; technology; expected operational environments; and, types of operations. No government can make a decision now and expect it is still optimal in fifteen years. Therefore, new plans are developed periodically that align with new predictions. These new predictions can be represented as selecting a new policy from a pre-computed set of solutions. However, governments must be very careful about when to continue; when to cancel; and when to switch manufacturing and purchasing orders. Changing policy directions can incur substantial financial penalties due to ramp-up and down costs; create periods of unbalanced forces during the switching period; require extra personnel training costs; etc. Therefore, a solution to this situation must be able to ensure that an optimal policy is maintained across objectives during the process of changing from one policy to another. This type of dynamic planning situation across multi-objective problems occurs frequently in real-world strategic decision making domains, such as government, business, manufacturing, etc. Hence, the development of robust solutions could support many decision makers.

\subsection{Wind farm control}
\label{sec:wind}

The design of traditional wind turbine control systems is typically focused on two objectives. On the one hand, a wind turbine needs to maximise its power production. On the other hand, maximising power output leads to higher fatigue loads (i.e., the stress induced on the turbine's components), and thus impacts their overall lifetime. Therefore, a trade-off needs to be made between power output and accumulated damage.

Single-turbine control and design has been well-explored in the literature \citep{review_maximum_power_point_tracking,review_turbine_control}. However, as multiple wind turbines are often geographically grouped into wind farms to reduce capital costs, the turbines become dependent on each other due to the wake effect. This effect occurs when upstream turbines extract energy from wind, leaving a cone of reduced available wind for downstream turbines, harming their productivity. One method to tackle this issue is through wake redirection control, where upstream turbines are purposely misaligned with the incoming wind vector in order to deflect wake away from the downstream turbines \citep{mats_journal_2020}. However, while misaligned rotors may lead to a higher farm-wide power production, it induces higher loads on the turbine's components. To tackle these non-linearities and complexities that originate from the wake effect, the use of data-driven optimisation methods is necessary to yield optimal wind farm control strategies.

Finding a good balance between power production and loads is challenging. The link between control actions, the high-dimensional load spectrum and future costs is still an open problem \citep{verstraeten2019fleetwide}. Therefore, although the relationship between control actions and maintenance costs is expected to be complex, a linear scalarisation of power production and loads is often employed (e.g., \citep{vandijk2016}), where the parameters are decided based on the expertise of operators. Preferably, the operators should receive a set of alternative control strategies to investigate, covering the entire spectrum of objectives ranging from load-focused to power-focused strategies.

\subsection{Other topics} 

In addition to the motivating examples discussed above, recent years have seen multi-objective learning and planning methods applied across a wide range of problem domains including: distributed computing \citep{qin2020energy,da2019multi}, drug and molecule design \citep{zhou2019optimization,horwood2020molecular}, cybersecurity \citep{sun2018pareto}, simulation \citep{ravichandran2018pedestrian}, job shop scheduling \citep{mendez2019multi}, cognitive radio networks \citep{messikh2018towards,raj2020survey}, satellite communications \citep{hu2020dynamic,ferreira2019reinforcement}, recommender systems \citep{lacerda2017multi}, power systems \citep{deng2018integrated,deng2020coordinated,wang2019multiobjective,mello2020cost}, building management \citep{zhang2019whole}, traffic management \citep{jin2019multi}, manufacturing \citep{govindaiah2019applying,lepenioti2020machine,dornheim2018multiobjective}, bidding and pricing \citep{yang2020motiac,krasheninnikova2019reinforcement}, education \citep{rowe2018toward}, and robotics \citep{huang2019learning}. The scope and variety of these applications supports our assertion that many important problems involve multiple objectives, and are best addressed using explicitly multi-objective methods.

\section{Problem setting}
\label{sec:bg}

First, let us introduce the basic multi-objective sequential decision problem. We formalise this as a \emph{multi-objective Markov decision process} (MOMDP). We note that more complex models exist, such as a multi-objective partially observable Markov decision process \citep{soh2011evolving,soh2011multi,wray2015multiPOMDP,nian2020dcrac} and multi-objective multi-agent systems \citep{radulescu2020survey}.  However, the MOMDP formalisation allows us to study many relevant aspects of multi-objective decision making problems, while also being simple to understand. We therefore use it as the basis for this article. In this section we will restrict discussion to single-agent MOMDPs and defer discussion of the more complex multi-agent situation until Section \ref{sec:momas_algos}.

A MOMDP is represented by the tuple $\langle S, A, T, \gamma, \mu, \mathbf{R} \rangle$, where: 
\begin{itemize}
    \item $S$ is the state space
    \item $A$ is the action space
    \item $T \colon S \times A \times S \to \left[ 0, 1 \right]$ is a probabilistic transition function
    \item $\gamma \in [0, 1)$ is a discount factor
    \item $\mu \colon S \to [0,1]$ is a probability distribution over initial states 
    \item $\mathbf{R} \colon S \times A \times S \to \mathbb{R}^d$ is a vector-valued reward function, specifying the immediate reward for each of the considered $d\ge2$ objectives
\end{itemize}

The crucial difference between a single-objective MDP and a MOMDP is the vector-valued reward function $\mathbf{R}$, which expresses a numeric feedback signal for each of the considered objectives. This means that the length of the reward vector is equal to the number of objectives. 

Like single-objective MDP, the state and action sets can in principle be discrete and finite.
However, in many real-world problems the state-space is infinite. This happens as soon as some of the state variables describing a state---such as the water levels in a lake (Section \ref{sec:water})---are continuous. Moreover, even if the state space is discrete, it often is too large to enumerate as states may be described using images, e.g., cameras in an autonomous car.
The action-space can also be infinite in size. For example, in wind farm control (see Section~\ref{sec:wind}), actions correspond to a specific rotor orientation with respect to the incoming wind direction. This again is a continuous value. Infinite state- and action-spaces make the problem considerably harder, and necessitate the use of function approximators to estimate policies and their (vector) values. 

\subsection{Policies and value functions}

In MOMDPs, the agent behaves according to a policy $\pi\in\Pi$, where $\Pi$ is the set of all possible (and allowed) policies. 
A policy is a mapping ${\pi}: S \times A \to \left[ 0, 1 \right]$, i.e., for any given state, an action is selected according to a certain probability distribution. 

The value function of a policy $\pi$ in a MOMDP is defined as: 
\begin{equation}
\mathbf{V}^{\pi} = \mathbb{E} \left[ \sum\limits^\infty_{k=0} \gamma^k \mathbf{r}_{k+1} \:|\: \pi, \mu \right],
\end{equation}
\noindent where $\mathbf{r}_{k+1} = \mathbf{R}(s_k, {a}_{k}, s_{k+1})$ is the reward received at timestep $k+1$. In contrast to single-objective MDPs, the value function is also vector-valued, $\bm{V}^{\pi} \in \mathbb{R}^d$. We can also define the value of a state $s$, for any timestep $t$, when $s_t=s$:
\begin{equation}
\label{eq:MO_state_value}
\mathbf{V}^{\pi}(s) = \mathbb{E} \left[ \sum\limits^\infty_{k=0} \gamma^k \mathbf{r}_{t+k+1} \:|\: \pi, s_t=s \right].
\end{equation}

In single-objective settings, the value functions offer a complete ordering over the policy space, i.e., for any two policies $\pi$ and $\pi'$, $V^{\pi}(s)$ will either be greater than, equal to, or lower than $V^{\pi'}(s)$. This implies that finding the optimal policy $\pi^*$ is equivalent to maximising the expected cumulative discounted reward. For a MOMDP this is not necessarily the case.

If we have access to a \emph{utility function} (also called a \emph{scalarisation function} in the literature) $u:\mathbb{R}^d\rightarrow\mathbb{R}$, mapping the multi-objective value of a policy to a scalar value,
\begin{equation}
    V_{u}^{\pi} = u(\mathbf{V}^\pi)  ,
\end{equation}
then this would give us a total ordering over policies and reduce the MOMDP to a single-objective decision making problem.
This however, is not always possible, feasible, or desirable as motivated in the introduction.  We illustrate this further in Section \ref{sec:design_factors:motivating_scenarios}. 

Thus, when dealing with multi-objective value functions (Equation~\eqref{eq:MO_state_value}), it is possible to encounter a situation in which $V_i^{\pi} > V_i^{\pi'}$ for objective $i$, while $V_j^{\pi} < V_j^{\pi'}$ for objective $j$. As a consequence, in MOMDPs, value functions only allow for a \emph{partial} ordering over the policy space, so determining the optimal policy is no longer possible without additional information on how to consider or prioritise the objectives to order the policies.   

Notice that the formulation of policies described in this section only allows for stationary policies, i.e., we condition only on the current state. While this may be sufficient for fully-observable, single-objective MDPs, \citet{white1982multi} demonstrate that for multi-objective tasks it may be beneficial to allow policies to be non-stationary with respect to the current state (i.e., also conditioned on other variables, such as the sum of previously received rewards). 

\subsection{Solution sets}
\label{sec:solsets}

In single-objective RL problems, there exist a unique optimal value $V^*$, and there can be multiple optimal policies $\pi^*$ that all have this value. The goal in single-objective RL is typically to learn an optimal policy.

In the multi-objective case however, without any additional information about the user's utility, there can now be multiple \emph{possibly optimal} value vectors $\mathbf{V}$.
We therefore need to reason about sets of possibly optimal value vectors and policies when thinking about solutions to MORL problems.
In the following, we introduce several useful definitions for possibly optimal policies and values.
We start by defining the most general set of solutions, i.e., the undominated set.
This is the set of policies and associated value vectors for which there is at least one utility function for which this policy is optimal (i.e., there is no other policy for this utility function that has strictly higher utility).

The concepts introduced in this section are defined in terms of policies. However, as each policy $\pi$ has an associated value vector $\mathbf{V}^\pi$, throughout the survey we often relate value vectors to these concepts when the context is clear.

\begin{Def}\label{def:u}
The undominated set, $U(\Pi)$, is the subset of all possible policies $\Pi$ and associated value vectors for which there exists a possible utility function $u$ whose the scalarised value is maximal:
\begin{equation}\label{eq:u}
	U(\Pi) = \left\{\pi \in \Pi\ \middle|\ \exists u, \forall \pi' \in \Pi : u(\mathbf{V}^{\pi}) \geq u(\mathbf{V}^{\pi'})\right\}.
\end{equation}
\end{Def}
However, the undominated set may well contain excess policies. That is, policies that are optimal for a given (set of) utility function(s), but where other policies exist that have optimal utility for that/those utility function(s) as well. In that case, we do not need to retain all policies to retain optimal utility. 

\begin{Def}\label{def:cs}
A set $CS(\Pi)$ is a \emph{coverage set} if it is a subset of $U(\Pi)$ and if, for every $u$, it contains a policy with maximal scalarised value, i.e.,
\begin{equation}\label{eq:cs}
	CS(\Pi) \subseteq U(\Pi) \wedge \left(\forall u, \exists \pi \in CS(\Pi), \forall{\pi' \in \Pi} : u(\mathbf{V}^{\pi}) \geq u(\mathbf{V}^{\pi'}) \right).
\end{equation}
\end{Def}

As mentioned, there generally does not exist a total ordering over the values of possible policies in a MORL problem. 
We can, however, again reason about sets of \emph{possibly optimal} policy values. 
For the most general case where $u$ is any (potentially unknown) monotonically increasing utility function (i.e., including non-linear functions), we define the set of undominated values as follows.

\begin{Def}\label{def:pf} 
If the utility function $u$ is any monotonically increasing function, then the \emph{Pareto Front (PF)} is the undominated set \citep{roijers2013survey}:
\begin{equation}
    PF(\Pi) = \{ 
    \pi \in \Pi\  |\ \nexists \pi'\in\Pi : \mathbf{V}^{\pi'} \succ_P \mathbf{V}^\pi
    \} ,
\end{equation}
where $\succ_{P}$ is the Pareto dominance relation,

\begin{equation}
    \mathbf{V}^\pi \succ_{P} \mathbf{V}^{\pi'} \Longleftrightarrow (\forall i: \mathbf{V}^\pi_i \ge \mathbf{V}^{\pi'}_i) \land (\exists i: \mathbf{V}^\pi_i > \mathbf{V}^{\pi'}_i).
\end{equation}
\end{Def}
In words, the Pareto Front is the set of non-dominated policies: for each policy in the Pareto Front, there exists no other policy with value that is equal or better in \emph{all} objectives.

Note that for the Pareto front this means we only need to retain one of the policies that have the same value vector. A set of policies whose value functions correspond to the PF is called a \emph{Pareto Coverage Set (PCS)}.

If the (a priori unknown) utility function is a positively-weighted linear sum, then the undominated set will be the policies corresponding to the convex hull (CH) of value functions $V^{\pi}$.

\begin{Def}\label{def:linscal}
A linear utility function computes the inner product of a weight vector ${\bf w}$ and a value vector ${\bf V}^\pi$
\begin{equation}\label{eq:linear}
	u({\bf V}^\pi)= {\bf w}^\top {\bf V}^\pi.
\end{equation}
Each element of ${\bf w}$ specifies how much one unit of value for the corresponding objective contributes to the scalarised value. The elements of the weight vector ${\bf w}$ are all positive real numbers and constrained to sum to 1.
\end{Def}
\begin{Def}\label{def:ch}
The \emph{convex hull (CH)} is the subset of $\Pi$ for which there exists a $\bf w$ (for a linear $u$) for which the linearly scalarised value is maximal, i.e., it is the undominated set for linear utility functions:
\begin{equation}
    CH(\Pi) = \{  \pi \in \Pi\  |\ \exists \mathbf{w}, \forall \pi' \in \Pi : \mathbf{w}^\top \mathbf{V}^\pi \ge \mathbf{w}^\top \mathbf{V}^{\pi'}\} .
\end{equation}
\end{Def}
In words, the convex hull is the set of policies that maximise the weighted sum over objectives for some weight vector $\mathbf{w}\in\mathbb{R}^d$. 

The Pareto Front and the Convex Hull often consist of infinitely many policies, especially when policies can be stochastic. 

However, coverage sets can often be significantly smaller.
This is particularly so for the \emph{convex coverage set}.

\begin{Def}\label{def:ccs}
A set $CCS(\Pi)$ is a \emph{convex coverage set} if it is a subset of $CH(\Pi)$ and if for every ${\bf w}$ it contains a policy whose linearly scalarised value is maximal, i.e., if:
\begin{equation}\label{eq:ccs}
	CCS(\Pi)\subseteq CH(\Pi) \wedge \left(\forall {\bf w}, \exists \pi \in CCS(\Pi), \forall{\pi' \in \Pi} : {\bf w}^\top {\bf V}^\pi \geq {\bf w}^\top {\bf V}^{\pi'}\right).
\end{equation}
\end{Def}

\noindent The CCS is not only important for linear utility functions. Specifically if we also allow stochastic policies in $\Pi$, a CCS is sufficient to construct a CS for all possible (non-linear) monotonically increasing utility functions as well, i.e., a PCS \citep{vamplew2009constructing}. 

For deterministic stationary policies, the difference between the $CH(\Pi)$ and a $CCS(\Pi)$ is often small. Therefore, the terms are often used interchangeably. The key difference however is stochastic policies. Specifically, if the space of deterministic policies is discrete (i.e., there are a finite number of states for which a finite number of actions can be chosen) then there is always a finite CCS, even if stochastic policies are allowed. In contrast, the CH is typically infinite in this case. This is especially important because, as we have already mentioned, this finite CCS can be used as a basis to construct every policy in a PCS. For more detailed information on these sets, and how they interact with deterministic/stochastic policy spaces, please refer to \citep{roijers2013survey}. 

The choice of solution set is key to the efficiency of the algorithms used to solve multi-objective problems. This is because we have to compute all the policies in these sets. When these sets become too large, we may not be able to compute them anymore, and we need to solicit more information on how to handle or prioritise the objectives. 
We consider that this optimisation process should be driven by the utility obtained by the user from a proposed solution which can be derived using the utility function. 
We introduce this perspective and approach in the following section.

\section{The utility-based approach}
\label{sec:utility_based}

Considering the user utility first is key to the successful application of any AI in decision problems. In multi-objective problems, it is especially important, as the properties of the user's utility may drastically alter the desired solution, what methods are available, and even---in some cases \citep{radulescu2020utility}---whether stable solutions even exist. Following the recent literature on multi-objective RL, we therefore support the \emph{utility-based} approach \citep{radulescu2020utility,radulescu2020survey,roijers2013survey,roijers2017multi,zintgraf2015quality}. 

The utility-based approach stands in contrast to the earlier axiomatic approach. In the axiomatic approach the optimal solution set to a multi-objective decision problem is assumed to be the Pareto front (see Definition~\ref{def:pf}).  However, this set is typically too large, and may be prohibitively expensive to retrieve. Furthermore, as \citet{vamplew2009constructing} have shown, if stochastic polices are allowed, a much smaller solution set suffices to construct a Pareto-front, i.e., we can use stochastic mixtures between the policies in the deterministic stationary convex coverage set (CCS), which is much easier to compute, and allows for algorithms that exploit the properties of the CCS to retrieve the optimal policies, such as outer loop methods \citep{roijersPhD} as discussed further in Section \ref{sec:multi-pol}. Moreover, in practical applications, a lot more might be known about the utility function of the user, due to domain knowledge. Using an axiomatic approach would make it difficult to exploit this knowledge, and a lot of time and effort might be spent on computing an approximate solution which contains solutions with very low utility for the user/deployment.

The utility-based approach aims to derive the optimal solution set from the  available knowledge about the utility function of the user, and which types of policies are allowed. This knowledge allows constraints to be placed on the solution set, reducing its size and thereby improving learning efficiency and making it easier for users or systems to select their preferred policy \citep{roijers2013survey}. The utility-based approach entails the following steps: 

\begin{enumerate}
    \item Collect all a priori available information regarding a user's utility.
    \item Decide which type of policies (e.g., stochastic or only deterministic) are allowed.
    \item Derive the optimal solution concept from the resulting information of the first two points. 
    \item Select or design a MORL algorithm that fits the solution concept. A variety of algorithms suited to differing solution concepts are reviewed in Section \ref{sec:survey_algorithms}.
    \item When multiple policies are required for the solution, design a method for the user to select the desired policy among these optimal policies. 
\end{enumerate}
\begin{figure}
    \centering
    \includegraphics[width=0.8\textwidth]{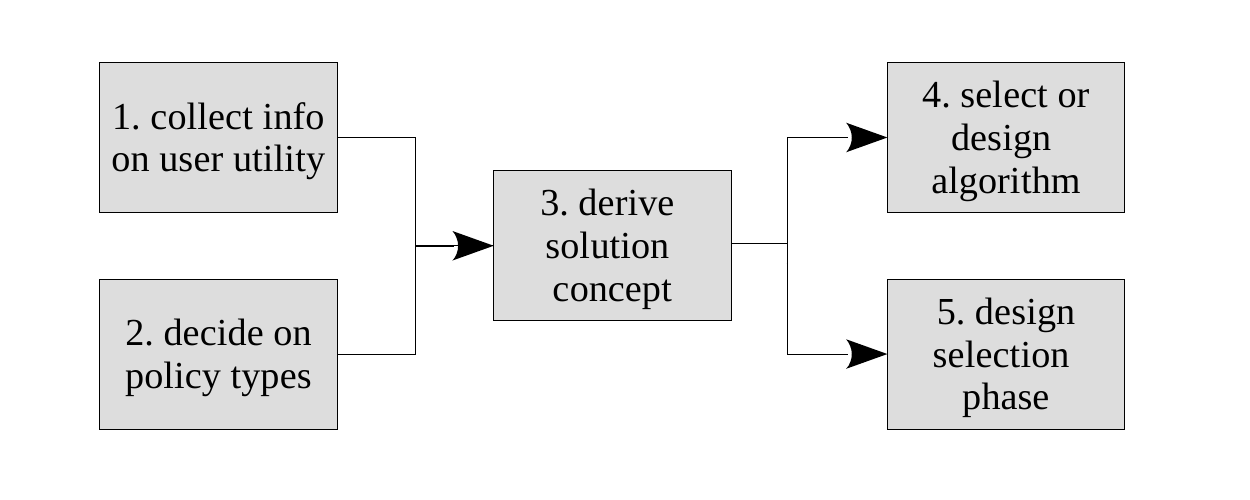}
    \caption{Work flow diagram for multi-objective reinforcement learning and planning.}
    \label{fig:wfd-morl}
\end{figure}
We note that some of these steps can be done in parallel, as illustrated in the work flow diagram in Figure \ref{fig:wfd-morl}. Specifically, it is possible to gather information on the user's utility and decide on which types of policies to allow simultaneously (Steps 1 and 2). However, Steps 1 and 2 need to be completed to be able to derive the solution concept (Step 3), which in turn needs to be completed before being able to select or design an appropriate algorithm (Step 4), and design how the user can select policies (Step 5).

In each of the steps to complete in this process, different factors will come into play. We will briefly discuss which factors must be considered at each step, while referring to later sections for a more detailed discussion. 

In Step 1, we aim to collect as much a priori available information about the user's preferences as possible. This information will help us determine the class of utility functions which we should employ. For example, if we know that all objectives correspond to units of goods that we need to buy or sell on an open market, the utility function will typically be linear (i.e., a sum of prices per unit, times the amount of units we need to buy and can sell). 

Another key distinction we have to make here concerns the application of the utility function for deriving the user's utility \citep{roijers2013survey,roijers2018esr,radulescu2020utility,radulescu2020survey}. Specifically, if the utility is derived from single outcomes of performing the policy, we need to apply the utility function to the returns, and then optimise the expected utility of the returns. This is the so-called Expected Scalarised Returns (ESR) criterion.  For example, in a medical treatment planning setting, the patients will derive their utility from their specific treatment outcomes. Conversely, if the utility is derived from the average returns over multiple runs we should take the expectation first, and optimise the utility of expected returns. This is called the Scalarised Expected Returns (SER) criterion. For a detailed discussion on whether to apply ESR or SER, please refer to Section~\ref{sec:seresr}.  

In Step 2, we need to decide what types of policies are allowed. This is important, as in contrast to single-objective problems, stochastic policies can be strictly better than deterministic policies \citep{vamplew2009constructing,white1980solution,white1982multi}. However, this does not mean that we should always allow them. For example, in a medical treatment planning setting, the patients would probably object to random selection of different medicines. Furthermore, we need to decide whether to allow non-stationary policies or not \citep{white1982multi}. For a detailed discussion on policy types, please refer to Section~\ref{sec:det-vs-stoch-policies}. 

Using the information from Step 1 and 2, we need to derive the appropriate solution concept (Step 3). For example, if the utility function is unknown at learning time, but known to be linear, any type of policy is allowed. We need a set of policies that contains at least one optimal policy for every possible set of linear weights. An example where this situation would arise would be where the linear weights correspond to fluctuating market prices of different commodities.

In Step 4, we need to either select an existing algorithm from the literature or design one that is suitable for the user's requirements. The choice of algorithm depends on the solution concept selected in Step 3; one of the main distinctions is between single-policy and multi-policy algorithms (see Section \ref{sec:morl_algos}). If the user utility function is completely known a priori and is not likely to change over time, a single-policy algorithm is appropriate. Conversely, if the utility function is unknown or subject to change a multi-policy algorithm is more suitable.

In Step 5, the goal is to help the user select a policy from a solution set produced by the algorithm selected in Step 4, that comes as close as possible to optimal user utility. This might be relatively straightforward if this set is small enough to show all possible policy value vectors to the user. If the set is large, or even continuous, more intricate methods are needed. For example, \citet{zintgraf2018ordered} use Gaussian processes to model the utility function, and use relative preferences queries posed to the user to train this model. Furthermore, they use targeted priors and additional (virtual) data to exploit the fact that utility functions in multi-objective decision problems are monotonic in all objectives.

Together, these steps form a complete pipeline to set up a multi-objective reinforcement learning or planning system.

\section{Factors influencing the design of multi-objective systems}
\label{sec:design_factors}

Multiple factors exist in multi-objective problem domains 
which do not need to be considered for single-objective problems, and these can have important implications for the design of a multi-objective agent. In this section, we identify and describe these factors, and explain the impact they may have on the design.

\subsection{Scenarios requiring a multi-objective approach}
\label{sec:design_factors:motivating_scenarios}
\begin{figure}
    \centering
    \includegraphics[width=0.9\textwidth]{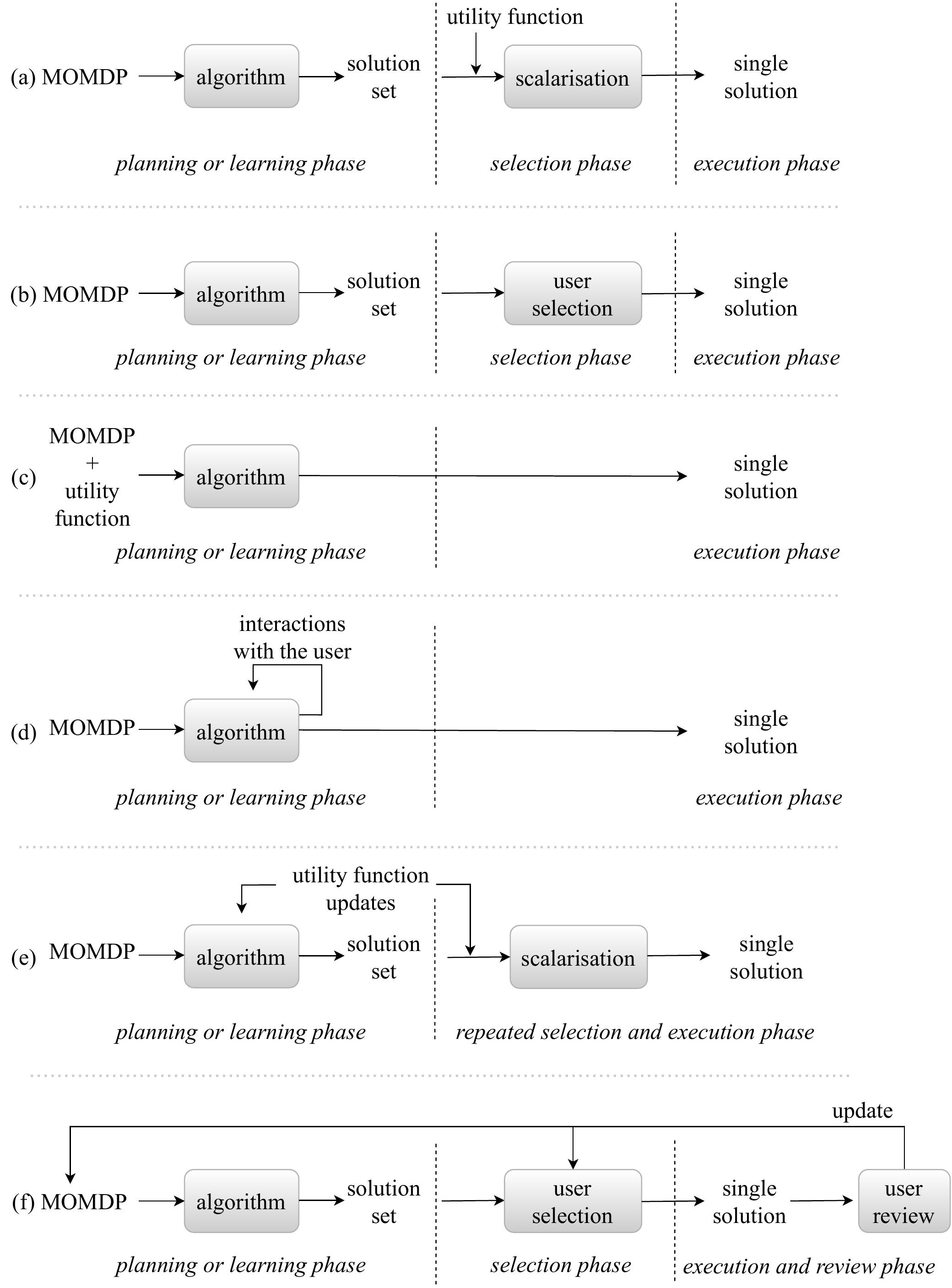}
    \caption{The six motivating scenarios for MOMDPs: (a) the unknown utility function scenario, (b) the decision support scenario, (c) the known utility function scenario, (d) the interactive decision support scenario, (e) the dynamic utility function scenario, and (f) the review and adjust scenario.}
    \label{fig:motivating_scenarios}
\end{figure}

Some researchers would argue that modelling problems as multi-objective is unnecessary and that all rewards can be represented as a single scalar signal. This implies that it is always possible to convert a MOMDP to a MDP. In order for this conversion to take place, an a priori scalarisation function is required. However, \citet{roijers2013survey} show that in certain situations it may be impossible, infeasible or undesirable to perform this conversion. \citet{roijers2013survey} present three scenarios in which this can occur as illustrated in (a), (b) and (c) in Figure~\ref{fig:motivating_scenarios}. Additionally, we propose three new motivating scenarios: the interactive decision support scenario (d), the dynamic utility function scenario (e), and the review and adjust scenario (f). Figure \ref{fig:motivating_scenarios} shows that each scenario consists of a planning or learning phase in which either a single policy or a solution set of multiple policies is found, and an execution phase in which a single policy is executed, and, in some scenarios, a selection phase in which the policy to be executed is selected.

In the \textbf{unknown utility function scenario} (a) \citep{radulescu2020utility}, a priori scalarisation is undesirable as the utility function is unknown at the time when planning or learning occurs. There is too much uncertainty around the utility that could be received. In this scenario it is preferable to compute a coverage set of policies so as to be able to respond quickly whenever more information is available. In the wind farm control example (Section~\ref{sec:wind}), there are two conflicting objectives. The goal is to maximise power output while minimising the required maintenance costs caused by the stress of operation. Specifying the exact preferences for these objectives is difficult since certain circumstances such as storms, the wake effect, and grid instability can affect the lifespan of turbine components. Since the link between these effects and preventive control measures is insufficiently understood, it is important to learn a set of optimal solutions.

In the \textbf{decision support scenario} (b), the user's preferences are unknown or difficult to specify. Working with this uncertainty makes it infeasible, if not impossible, to use a priori scalarisation as the user's utility function is unknown. The decision support scenario is almost identical to the unknown utility function scenario. The only difference is during the selection phase, where a set of policies is presented to a user who selects a policy based on their preferences. In the water management example (Section \ref{sec:water}), the optimal solution for managing a water reservoir depends on many stakeholders and their multiple conflicting objectives. Each stakeholder has their own preferences as to how the water should be managed, with each objective having an effect on different aspects of the businesses operating around the lake as well as the livelihood of those living nearby. Capturing accurate preferences for all stakeholders while taking into account the trade-offs across all objectives would be difficult, if not impossible. Instead, it would be better to learn a set of optimal policies and then make decisions regarding what policy to follow once a collective decision can be made by a local council or government.

The difference between the two scenarios above lies in the selection phase. The first scenario includes a utility revelation step where the utility function is made explicit. In the second scenario on the other hand, the decision relies on the user(s), and the utility function remains implicit in the decision taken.  As defining a utility function explicitly is hard (if not infeasible), user selection typically employs the decision support scenario (b).

In the \textbf{known utility function scenario} (c), the user's preferences are known. Working with known preferences, we can assume the user's utility function are known at the time of learning or planning, making scalarisation both possible and feasible. However, it can still be undesirable to do so because performing a priori scalarisation can lead to an intractable problem \citep{roijers2013survey,radulescu2020utility}. In the wind farm control example (Section \ref{sec:wind}), based on their preferences a user may want to maximise power output while minimising the stress on the turbine's components. Since the user's preferences are known, it is possible to learn a single policy that optimises the user's preferences.

In the \textbf{interactive decision support scenario} (d), the agent has to learn about both the preferences of the user and the environment \citep{roijers2018interative}. Applying a priori scalarisation in this scenario can be both undesirable and infeasible as the utility function of the user may be unknown or may be uncertain. During learning, the agent can elicit preferences from the user, removing uncertainty about the user's utility function. In the planning a journey example (Section \ref{sec:journey}), a user may not be able to accurately specify their preferences. While cost and travel time are preferences that may be easily specified, objectives such as comfort and reliability may be difficult to specify. When planning a journey other trade-offs such as taking a direct route over switching trains or having a lay over may have an impact on the user's utility. At various times during the learning phase a user could be presented with different potential solutions and rank the solutions in order of preference. This will enable the system to get a more accurate representation of the users preferences and learn an optimal solution for the users.

In the \textbf{dynamic utility function scenario} (e), the user's preferences for certain objectives change over time \citep{natarajan2005dynamic}. In this scenario applying a priori scalarisation would be undesirable. Given that a user's preferences can change over time, it would be optimal for the algorithm to learn a finite number of policies over time and choose an appropriate non-dominated policy for any utility function and improve upon it. Although there is an infinite amount of utility functions, they can be covered by a finite number of policies \citep{natarajan2005dynamic}. In the military purchasing example (Section \ref{sec:military}), current governments must make decisions about military purchasing, but as governments change over time so do the preferences of each government towards military spending. Using a system that can learn optimal policies for changing preferences is the desired approach for this example. While it would be possible to learn a single policy for the initial utility function and then dynamically adapt this as the user's utility function changes, this would incur a period of sub-optimal behaviour as the agent adapts, which need not occur if the agent has learned in advance a suitable set of solutions. When governments change it is crucial that optimal policies are still followed despite the change in preferences. 

In the \textbf{review and adjust scenario} (f), a user may be uncertain about their preferences over objectives and their preferences over objectives could change over time. Applying a priori scalarisation in this scenario is unfeasible as there is too much uncertainty around the utility function of the user. In this scenario, learning a coverage set of policies is optimal. Once a coverage set has been learned a user can then select the policy which accurately reflects their preferences. Before execution, the user can review their chosen solution. If the user's preferences have changed, the user can adjust their selected solution to accurately reflect their updated preferences. 

The review process can also update the MOMDP which can alter the set of solutions learned. This may for example occur when a new objective is identified, that was previously missed. For example, imagine an agent is used to control traffic in a part of a city. Initially, the pollution levels are seen as a single objective. However, after inspecting a map of the pollution levels resulting from the policies, it turns out pollution levels around a school is relatively high, while a school is actually an area where it ought to be low. In such a case, the pollution objective should be refined, i.e., split into two objectives: one overall, and one for sensitive/key areas such as schools (and e.g., hospitals).

In the planning a journey example from Section \ref{sec:journey}, a user may not be certain about their preferences for comfort and cost, among other objectives. In this scenario a set of optimal solutions is learned and the user selects the solution which accurately reflects their preferences. However, before execution if the user's preferences have changed, the user can review their chosen solution and select an alternative solution which accurately reflects their updated preferences. The system can then be updated to reflect the newly obtained information about the user's preferences.

 \subsection{Problem taxonomy}

\citet{roijers2013survey} outline a problem taxonomy which discusses what constitutes an optimal solution for a MOMDP. The taxonomy is based on the utility based approach, where the agent's ultimate goal is to maximise user utility \citep{radulescu2020utility}. In the previous section, we highlighted three new motivating scenarios and we have updated the problem taxonomy diagram from \citet{roijers2013survey} to include the extended scenarios. The taxonomy in Table \ref{table:taxonomy_table} outlines how each factor can lead to different solution concepts. It is important to carefully consider each factor in the taxonomy before choosing a solution concept.
The factors of the problem taxonomy are covered extensively in \citep{roijers2013survey}, but we will briefly outline each factor below. 

\begin{table}[h]
\centering
\begin{tabular}{l|l|l|l|l|}

\multicolumn{1}{l}{} & \multicolumn{2}{l}{\makecell{\emph{single policy} \\\emph{(known utility function,} \\\emph{interactive decision support)} \\\\}}  & \multicolumn{2}{l}{\makecell{\emph{multiple policies} \\\emph{(unknown utility function}, \\\emph{decision support}, \\\emph{dynamic utility function}, \\\emph{review and adjust}) \\\\}}                                                   \\ 
\multicolumn{1}{l}{} &\multicolumn{1}{l}{\makecell[l]{deterministic}} &\multicolumn{1}{l}{\makecell[l]{stochastic}} &\multicolumn{1}{l}{\makecell[l]{deterministic}} &\multicolumn{1}{l}{\makecell[l]{stochastic}}    \\ \cline{2-5}

\makecell[l]{linear \\scalarisation} &   \multicolumn{2}{l|}{\makecell[l]{one deterministic \\stationary policy \\\\}}         & \multicolumn{2}{l|}{\makecell[l]{convex coverage set of \\deterministic stationary policies \\\\}}   \\ \cline{2-5} 
\makecell[l]{monotonically \\increasing \\utility function}   &\makecell[l]{one \\deterministic \\non-stationary \\policy \\\\}         &   \makecell[l]{one mixture \\policy of two \\or more \\deterministic \\stationary \\policies \\\\}        
&\makecell[l]{Pareto \\coverage set of \\deterministic \\non-stationary \\policies \\\\}      & \makecell[l]{convex \\coverage set of \\deterministic \\stationary \\policies \\\\}       
\\ \cline{2-5} 

\end{tabular}
\vspace{0.2cm}
\caption{Taxonomy of multi-objective decision problems and corresponding solution sets. See Section \ref{sec:solsets} for the definitions of the solution sets.}
\label{table:taxonomy_table}
\end{table}

\subsubsection{Single versus multiple policies}
Whether or not an algorithm learns a single or multiple policies depends on which of the motivating scenarios holds from Section \ref{sec:design_factors:motivating_scenarios}. For example, in the unknown utility function and decision support scenarios the agent needs to learn multiple policies. In both of these scenarios the utility function of the user is unknown at the time of learning or planning, and therefore the agent must return a set of optimal policies. 
In the known utility function scenario the user's utility function is known at the time of learning or planning and therefore returning multiple policies is not necessary.

In the planning a journey example (Section \ref{sec:journey}), a user may or may not know their exact preferences about getting to their destination. A user may be unsure about how they would like to get to their destination or how much they are willing to spend on the journey. In this case we are in the unknown utility function scenario (a) \citep{roijers2013survey} and learning a coverage set of policies is required. In contrast, a user may want to arrive to their destination at a specific time using a specific mode of transport, and may have a fixed budget. Since the user's preferences are known we are in the known utility function scenario (c) \citep{roijers2013survey} and a single policy which represents the user's preferences can be learned.
\subsubsection{Linear versus monotonically increasing utility functions}
 The nature of the utility function has a significant role to play in what constitutes an optimal solution in a MOMDP and which of the motivating scenarios holds. When the utility function is linear the weighted sum for each value of the objectives is computed. In the known utility function scenario, the utility function is known at the time of learning or planning. The utility function can be applied to each reward vector in the MOMDP and an optimal solution can be found. But linear utility functions may not be suitable when trying to express a user's preferences. If a user's preferences are non-linear, a linear utility function is unable to accurately represent these preferences.
 
A monotonically increasing utility function adheres to the constraint that if a policy increases for one or more of its objectives without decreasing any of the objectives, the scalarised value also increases. A monotonically increasing utility function is able to represents both linear (with non-zero positive weights) and non-linear user preferences. For example, in the unknown utility function scenario, the agent must learn a set of policies. When the utility function is unknown at the time of learning or planning, Pareto dominance can be used to determine a set of non-dominated solutions. Since the utility function is monotonically increasing, policies that are Pareto dominant will be preferred by the user.

In the wind farm control example (Section \ref{sec:wind}) there are two  objectives: to maximise power and to reduce fatigue loads on the turbine. In the real world, a user will likely have non-linear preferences over these objectives. If these preferences are known (known utility function scenario) at the time of learning or planning it is crucial they are represented using a non-linear utility function. If the user's preferences are represented using a linear utility function a sub-optimal solution will be learned. A linear utility function cannot accurately represent non-linear preferences \citep{roijers2013survey}. In this case, learning a sub-optimal solution could negatively impact a wind turbines performance. If a wind turbine is not operating optimally, stress on the turbine's components would be increased, which impacts the components lifespan and increases maintenance costs. However, if a non-linear utility function is used to represent the user's preferences then it is possible to learn an optimal solution.

\subsubsection{Deterministic versus stochastic policies}\label{sec:det-vs-stoch-policies}
Whether to restrict the agent to policies that are deterministic or to allow stochastic policies has a significant impact on what an optimal solution is in a MOMDP. When the utility function is linear, we can translate the MOMDP to a single-objective MDP. In an MDP, only deterministic stationary policies apply as the optimal obtainable value is reachable with deterministic stationary policies. This is true for all linear utility functions.
But when the utility function is monotonically increasing and non-linear the situation is much more complex.

For example, in the known utility function scenario, the utility function is known during the learning or planning phase. If the utility function is a linear representation of a user's preferences, we can then translate the MOMDP to a single-objective MDP where only deterministic stationary policies hold. As another example, in the unknown utility function scenario where the utility function is assumed to be non-linear and only deterministic policies are allowed a coverage set of Pareto dominant policies must be learned. In this scenario, non-stationary policies can Pareto dominate stationary policies \citep{white1982multi}, therefore the Pareto coverage set must include non-stationary policies \citep{roijers2013survey}.

In the water management example (Section \ref{sec:water}) there are certain scenarios where stochastic policies should not be considered whatsoever. A stochastic policy where there is a chance the dam gates are opened and all the water in the reservoir is drained should not be considered, even if other outcomes of the policy are optimal. This stochastic policy would have catastrophic outcomes for the nearby town. If the utility function is non-linear and known at the time of learning or planning (known utility function scenario (c)) it would be optimal to learn one deterministic non-stationary policy. In this case, devastating outcomes like the scenario already mentioned would be avoided.

\subsection{Scalarised expected returns and expected scalarised returns}
\label{sec:seresr}

In contrast to single-objective reinforcement learning, in multi-objective reinforcement learning (MORL) different optimality criteria exist \citep{roijers2013survey}. Optimising under each criterion can lead to significantly different policies being learned \citep{radulescu2020utility}, where the criterion chosen for optimisation depends on how the policies are used in practice. The two optimisation criteria are known as the scalarised expected returns (SER) and the expected scalarised returns (ESR). 

The SER criterion is the most commonly used optimisation criterion in multi-objective RL and planning \citep{roijers15computing}. The SER criterion is calculated by first computing the expected vector returns of a policy and then applying the utility function to this expectation, 
\begin{equation}
    V_{u}^{\pi} = u\left(\mathbb{E} \left[ \sum\limits^\infty_{i=0} \gamma^t {\bf r}_i \:\middle|\: \pi, s_0 \right]\right).
    \label{eqn:ser}
\end{equation}
For SER the utility of a user is derived from multiple executions of a policy. SER is the optimal optimisation criterion in scenarios where the user is concerned about achieving an optimal utility over time. For SER, a coverage set is defined as a set of optimal solutions for all possible utility functions. 

However, many scenarios exist where only a single execution of a policy may be relevant to a user. In scenarios where a single execution of a policy is used to derive the utility of a user, optimising under the ESR criterion is optimal \citep{roijers2018esr}. For example, in a medical setting a patient may have only one opportunity to select a treatment. Under the ESR criterion the utility function is applied to the returns and the expectation is then computed,
\begin{equation}
    V_{u}^{\pi} = \mathbb{E} \left[ u\left( \sum\limits^\infty_{i=0} \gamma^t {\bf r}_i \right) \:\middle|\: \pi, s_0 \right].
    \label{eqn:esr}
\end{equation}
For a linear utility function there is no difference in the policies learned for SER and ESR. However, for a non-linear utility function the policies learned under SER and ESR are significantly different \citep{radulescu2020survey}.
 Many RL methods cannot be combined with the ESR criterion because non-linear utility functions in MOMDPs do not distribute across the sum of immediate and future returns which invalidates the Bellman equation \citep{roijers2018esr},
\begin{equation}
\max_\pi \mathbb{E} \left[u\left({\bf R}_t^- + \sum_{i=t}^{\infty} \gamma^i {\bf r}_i\right)\ \middle|\ \pi, s_t \right] \not= u({\bf R}_t^-) + \max_\pi \mathbb{E}\left[  u\left( \sum_{i=t}^{\infty} \gamma^i {\bf r}_i\right)\ \middle|\ \pi, s_t \right],
\end{equation}
where $\bf{R^{-}_{t}}$ $=$ $\sum_{i=0}^{t - 1} \gamma^i {\bf r}_i$.
ESR is the most commonly used optimality criterion in the game theory literature and literature on multi-objective games \citep{radulescu2020survey}. However, the ESR criterion has been extensively understudied in the RL literature. To study the ESR criterion further it is essential that new methods are formulated. It is important to note, for the ESR criterion a coverage set has yet to be defined and this is an open area for research.

In the planning a journey example (Section \ref{sec:journey}), we can consider a known non-linear utility function for a user planning their daily commute to work. If the user's employer is flexible about what time the user can start work, then the user can sometimes be late as long as the user is usually on time. Under the SER criterion a policy is optimised on the duration of journey objective. Since this policy is executed everyday it is acceptable for the user to be late some days because the days when the user is early compensate. However, if the user's employer is strict and requires the user to be on time each day or be subject to a fine, it is crucial that the user can plan a daily journey where they arrive on time. In this case optimising under the ESR criterion is optimal since every policy execution must ensure that the user arrives to work on time.

\section{The relationship with other problems}
\label{sec:relationships}

Parallels exist between some aspects of multi-objective sequential decision-making tasks and the classes of problems considered by other areas of reinforcement learning and planning research. In this section we identify some of the key areas of overlap between fields where we believe there is potential for beneficial exchange of ideas and techniques. We also pinpoint several pitfalls where the application of methods to multi-objective problems may not be consistent with the utility-based paradigm.

\subsection{Partially observable MDPs}
\label{sec:pomdp}

A key observation made already in the 1980s, is that if one assumes linear utility functions, POMDPs are a superclass of MOMDPs \citep{white1980solution}. To see this, imagine there would be a ``true objective'' and the linear weights of the utility function would form a ``belief'' over what the true objective would be. This is a special type of POMDP, where there will never be any observations concerning what the ``true objective'' is -- because after all, it does not actually exist. 

Of course multi-objective problems and partially observable problems have significantly different interpretations. However, the fact that POMDPs form a superclass of multi-objective MDPs under linear utility has important consequences for researchers and practitioners alike. Firstly, a lot of theoretical properties are inherited from POMDPs. This means that a lot of theorems do not have to be proven anew for MOMDPs under linear utility. So if you are wondering whether a certain property holds, it is prudent to consult the POMDP literature as well. Secondly, it means that methods that have been invented originally for POMDPs, can often be adapted for usage in MOMDPs \citep{roijersPhD}. While doing so, it is key to note that the number of objectives in a MOMDP correspond to the number of states in a POMDP (i.e., the dimensionality of the belief- and $\alpha$-vectors) \citep{roijers15computing}. This means that methods that did not work well in a POMDP context because they scale poorly in the number of states, might be very useful in a MOMDP context. A good example of this is Optimistic Linear Support (OLS) \citep{mossalam2016multi,roijersPhD,roijers15computing}, which was based on Cheng's linear support for POMDPs \citep{cheng1988algorithms}. Finally, it might mean that some algorithmic improvements may be applicable to both MOMDPs and POMDPs (such as \citep{roijers2018bootstrapping}).

\subsection{Multi-objective as multi-agent problems}

\paragraph{Objectives are not agents.} Some papers---in our opinion abusively---cast single-agent multi-objective problems as multi-agent problems, with each agent representing a competing objective \citep{li2019urban,mendez2019multi}. Then, either through voting rules \citep{tozer2017many} or (Nash) equilibria \citep{duan2014multi,economides1991multi,lee2012multi,li2012application}, a policy is selected. This mechanism however, has no guarantees with respect to user utility. It is unclear whether this ``compromise solution'' represents a desired trade-off or not. Specifically, the concepts of voting rules and Nash equilibria have been designed to find trade-offs between the individual utilities of agents. This is different from trade-offs for an individual agent between objectives, as objectives can be more or less important and may have non-linear interactions in the utility function. A voting rule or Nash equilibrium is not able to capture such subtleties and can therefore function as no more than an unfounded heuristic. In fact it is well-known that Nash equilibria can be Pareto-dominated \citep{Cohen9724,dubey1990inefficiency}.

\paragraph{But altruistic agents can see other agents as objectives.}
On the other hand, if we consider an agent that is explicitly altruistic, i.e., it cares about the other agents in its environment, such an agent could see the utility of these other agents as objectives, and therefore this should be modelled as a multi-objective problem. 
As such, it is also possible to consider varying levels of altruism. \citet{aoki2004distributed} for example, consider a multi-stage flow system with multiple agents. Each agent is a service centre, and is represented as a different objective by the other agents. They use a distributed reinforcement learning framework and propose a bi-directional decision making mechanism to address the resulting multi-objective problem. 

Modelling other agents as objectives enables explicitly imposing fairness between these objectives, i.e., the utilities of the agents. A loose condition for fairness is that a joint policy $\boldsymbol{\pi}$ is not so-called Lorenz dominated. Lorenz domination is based on the so-called Lorenz vector \citep{perny2013approximation}. 
    The Lorenz vector ${\bf L}({\bf V}^{\boldsymbol{\pi}})$ of a vector ${\bf V}^{\boldsymbol{\pi}}$ is defined as:
    \begin{equation}
    \left( v_{(1)}, ~ v_{(1)} + v_{(2)}, ~ ... ~ ,~ \sum_{i=1}^N v_{(i)} \right), 
    \end{equation}
    where $v_{(1)} \leq v_{(2)} \leq ... \leq v_{(N)}$ correspond to the values in the vector ${\bf V}^{\boldsymbol{\pi}}$ sorted in increasing order.
Imposing that the Lorenz vector of a policy is undominated leads to the Lorenz optimal set as coverage set.
    A vector ${\bf V}^{\boldsymbol{\pi}}$ Lorenz dominates ($\succ_L$) a vector ${\bf V}^{\boldsymbol{\pi'}}$ when:
    \begin{equation}
    {\bf V}^{\boldsymbol{\pi}} \succ_L {\bf V}^{\boldsymbol{\pi'}} \Leftrightarrow {\bf L}({\bf V}^{\boldsymbol{\pi}}) \succ_P {\bf L}({\bf V}^{\boldsymbol{\pi'}}),
    \end{equation}
    i.e., when the Lorenz vector of  ${\bf V}^{\boldsymbol{\pi}}$ Pareto dominates the Lorenz vector of  ${\bf V}^{\boldsymbol{\pi'}}$.
A Lorenz optimal set can then be input for a negotiation of which policy to execute in practice \citep{radulescu2020survey}.  Lorenz optimal sets have been studied in the context of different problem domains \citep{galand2015exact,moghaddam2011lorenz,perny2013approximation}. 

In short, objectives do not typically correspond to the interests of single agents as agents will care about multiple objectives. However, altruistic agents may see the interests of other agents as objectives, and therefore aim to come up with fair solutions. 

\subsection{Multi- and auxiliary task RL}

A highly related problem that has recently gained traction in the RL literature is that of \emph{auxiliary tasks} and  \emph{multi-task RL}. For example, \citet{schaul2015universal} define multiple goals, which are typically a subset of the states. They then learn a universal value function approximation (UVFA) network, that learns a value with respect to these different goals. From a MORL perspective, UVFA is thus an instance of MORL, with the restriction that goals are closely associated with states, and that the utility function may only select one of these goals to be \emph{the goal} at the moment. To move to a more general  MORL setting, a goal should be generalised to a specific a priori known (parameterised) utility function and as such there is no clear relation between the goal (i.e., the importance of each objective) and the state. This would be an issue in multi-policy settings as (a) it is not clear how many specific utility functions would be needed, and (b) non-linear utility functions would not be supported. 
For the dynamic weights setting with linear utility functions, UVFA has been adapted to the MORL setting as a baseline algorithm \citep{abels2019dynamic}, and shown to perform worse than specific MORL algorithms, but better than more naive baselines. 

In their work on successor features (SF), \citet{barreto2017successor} decompose
a scalar reward into a product of state features and task
weights to enable transfer learning between tasks. Again, we observe that successor features are in fact a subclass of multi-objective problems with linear weights, i.e., where objectives can be associated with desirable state-features. Universal Successor Features Approximators \citep{borsa2018universal} and Universal Successor Representations \citep{ma2018universal} combine the benefits of SF and UVFA to further
generalise across goals. It is important to note though that while state features and task/goals weights are analogous to the multi-objective
reward and linear weight vectors, in MORL the decomposition between reward and weight vectors is typically given rather than learnt. This is because successor features are not observing the individual objectives and are only provided with a scalar reward function. One might think that this would make SF more widely applicable than MORL. However, it also restricts the usage of such methods to things that can be inferred from state features. But, more importantly, scalar reward functions are often engineered on the basis of real events, multiple sensor inputs, and endlessly tweaked on the basis of the actual objectives of the users and designers, as we discussed in Section~\ref{sec:introduction}. Hence, using successor features instead of MORL, would in many real-world problems come down to throwing away information first in order to construct a scalar reward function, to later partially infer it back from data. This is of course sub-optimal, and should be avoided if possible.

\subsection{Human-aligned agents}

As AI systems are increasingly being applied to important real-world tasks, interest has grown in ensuring that the behaviour of autonomous systems is aligned with our own objectives, so as to avoid harmful outcomes either at a general level or with regards to specific individuals.  Research within this field focuses on ensuring that the decisions and behaviour of autonomous agents are safe, trustworthy,aligned interpretable, fair and unbiased. As these add additional considerations beyond maximising the agent's primary reward, there is a clear link to multi-objective approaches.

Strong parallels exist between multi-objective decision making and risk-sensitive or safety-aware decision making. An agent making decisions in the context of uncertain risks must aim not just to maximise its expected reward, but also to account for some measure of risk. This measure may be based on the variance of the reward, the worst-case outcome or the probability of entering known error states \citep{garcia2015comprehensive}. As with the multi-objective methods discussed in this paper, the choice of optimal action for a risk-aware agent will be based on combining together the expected reward and risk measures for each action using some form of utility function. Therefore it is not surprising that several authors have framed safe reinforcement learning as a multi-objective problem. In \citep{geibel2005risk}, \citep{geibel2006reinforcement} and \citep{horie2019multi}, MORL was applied to develop risk-aware agents, where the risk-related reward is based on the probability of the agent visiting an error state. Meanwhile, Elfwing and Seymour \citep{elfwing2017parallel} argue, based on biological evidence, that computational agents may behave more safely if they learn separate values for rewards and punishments.

As well as a growing interest in safe AI, recent years have also seen an increasing focus on the issues of explainability and interpretability of autonomous systems, as these factors are important for building trust with human users, and in ensuring transparency and lack of bias. It has been argued that a reward which has been decomposed from a scalar into its component terms provides benefits from the perspective of explaining decisions \citep{juozapaitis2019explainable}, and so several recent papers have explored multi-objective approaches to explainable and interpretable RL agents \citep{noothigattu2018interpretable,zhan2019relationship,cruz2019memory}.

In many applications it is also important to ensure that the actions of an agent are fair with regards to multiple stakeholders---a solution which is optimal for many members of society but which significantly disadvantages a sub-set of the population may well be unacceptable. In some contexts this may involve the development of multi-agent systems as in our earlier discussion of altruistic agents. In other situations, a single agent may be considering multiple objectives where each objective corresponds to the desires of a particular individual stakeholder, or where each stakeholder may have different preferences over these objectives and the agent must balance the utility obtained by each stakeholder. Multi-objective approaches to fairness have been explored at an abstract level \citep{siddiquelearning}, and also within specific applications \citep{hu2020dynamic,perez2010multi}.

In short, following \citet{vamplew2018,vamplew2021potential}, we argue that multi-objective agents provide a suitable mechanism for developing human-aligned artificial intelligence, addressing safety constraints as well as other alignment issues such as ethics or legal restrictions.

\section{Survey of multi-objective reinforcement learning and planning algorithms}
\label{sec:survey_algorithms}
In this section we review the state-of-the-art in algorithms for multi-objective planning and reinforcement learning, relating these algorithms back to the design factors identified in Section~\ref{sec:design_factors}. The aim is to aid in identifying which extant algorithms may be best suited for a particular application, based on the properties of that application.

\subsection{Multi-objective planning algorithms}

Research on planning approaches to MOMDPs has been established for much longer than work on reinforcement learning approaches, dating back to at least the early 1980s \citep{white1980solution,thomas1982constrained,white1982multi}. White and Kim \citep{white1982multi} adapted dynamic programming to develop an algorithm for finding Pareto set policies for infinite horizon discounted MOMDPs. However, as identified by \citet{wiering2007computing}, that approach has issues of computational feasibility and finds policies which are non-stationary. To address this, they developed the CON-MODP algorithm which invokes a consistency operator to ensure the stationarity of policies.

\citet{bryce2007probabilistic} demonstrated by example that, in the context of MOMDPs with stochastic state transitions, agents which aim to maximise the SER cannot rely on localised decision-making. The information available at any given state is insufficient to determine the optimal action under the SER formulation, and the agent must also take into account the actions which will be selected, and rewards which will be received, at all other states of the MOMDP. They develop the Multi-objective Looping $AO^*$ ($MOLAO^*$) algorithm to address this issue.

The Convex Hull Value Iteration (CHVI) algorithm \citep{barrett2008learning} is amongst the most widely-cited works on MOMDP planning. Although it is frequently incorrectly described as a MORL method, it in fact extends Bellman's value iteration algorithm to estimate and store the convex hull of future rewards for each state-action pair. This allows CHVI to identify the coverage set of policies, but only under the assumption that the utility function is linear. Because of the linear utility function, CHVI is akin to planning in POMDPs (see also the relation with POMDPs in Section \ref{sec:pomdp}). This has recently been shown in a paper that improves both CHVI and POMDP value iteration methods by reusing information across linear programs in subsequent iterations of these methods \citep{roijers2018bootstrapping}.

Other planning methods have considered the possibility of specific non-linear definitions of utility. Perny and Wang \citep{perny2010finding} address the task of finding the single optimal policy given the goal of minimising the distance between the reward vector received and a target reference point in objective space. They show that the non-linear nature of this utility prevents direct adaptation of methods like dynamic programming which are based on the Bellman equation, and instead develop a non-linear programming solution for this task. Meanwhile, \citet{wray2015multiMDP} identify Lexicographic MDPs as a specific subset of MOMDPs, where there is a specified ordering over objectives. They develop methods based on value-iteration for solving such tasks, allowing the ordering of objectives to be state-dependent and incorporating the concept of slack, which allows some degree of loss in the primary objective in order to obtain gains in secondary objectives.  This approach has also been extended to POMDPs \citep{wray2015multiPOMDP}. 

\subsection{Multi-objective reinforcement learning algorithms}
\label{sec:morl_algos}

\subsubsection{Stateless/bandit algorithms}

Algorithms designed for the multi-armed bandit (MAB) domain endeavour to follow an optimal exploration/exploitation strategy for selecting between different actions (arms), so as to minimise the regret (the loss in reward from not selecting the, initially unknown, optimal action on every time-step). Several papers have examined the extension of MAB algorithms to multi-objective tasks, often by adopting the concept of multi-objective regret in which the agent aims to minimise the number of Pareto-dominated actions which are performed.

Several multi-objective variations to the well-known UCB1 algorithm are compared in \citep{drugan2013designing}, including linear and Chebyshev scalarisations, as well as a version based on Pareto dominance. The empirical results show that Pareto UCB1 outperforms the scalarised versions. Later, \citet{yahyaa2014knowledge} demonstrated that a Pareto-based variant of the knowledge gradient algorithm could lead to further improvements in performance over Pareto UCB1.

Other work has examined multi-objective extensions to specialised forms of bandits. Van Moffaert and Now\'e \citep{van2014multi} consider a multi-objective form of the $\chi$-armed bandit, in which the set of arms is a measurable (potentially infinitely large) set of arms. They propose a modified form of the Hierarchical Optimistic Optimization (HOO) algorithm for this class of bandits. Likewise, Lacerda \citep{lacerda2017multi} examines multi-objective extensions of ranked bandits, in which the agent produces a ranking of arms rather than a single choice at each time-step. More recently, \citet{turgay2018multi} extended the contextual MAB model to incorporate multiple objectives. Unlike conventional MABs, a contextual bandit incorporates some additional state or side-information, and so represents a compromise between stateless bandits and full-blown RL scenarios. Their Pareto Contextual Zooming (PCZ) algorithm aims to minimise the Pareto regret while also maintaining a fair distribution over the Pareto-optimal arms \citep{bouneffouf2020survey}.

\subsubsection{Single-policy algorithms}

Perhaps the simplest and most widely-adopted approach to MORL is to extend existing single-objective model-free value-based methods, such as Q-learning, to handle multiple objectives. This extension requires two changes to the learning algorithm, i.e., the agent must store Q-values as vectors rather than scalars, and the scalarisation function designed to match the user's utility function must be used to identify the greedy-action to perform in any given state. This approach naturally gives rise to single-policy solutions to the multi-objective problem, as the underlying single-objective methods are designed to produce a single optimal solution.

Many applications of this approach have used a linear scalarisation function, either weighted or unweighted \citep{aissani2008efficient,shabani2009incorporating,guo2009reinforcement,perez2009responsive}. This is equivalent to transforming the MOMDP into a corresponding MDP, and so existing proofs of convergence apply \citep{roijers2013survey}. In some domains this will also be a suitable representation of the user's underlying utility (for example, in problems where the objectives are naturally expressed in monetary terms). However, in many cases this linear function will be inadequate to represent the user's true utility \citep{vamplew2008limitations}. Therefore it will often be preferable to use a non-linear function instead \citep{gabor1998multi,van2013hypervolume,van2013scalarized,issabekov2012empirical}. Nevertheless, this violates the assumption of additive returns in the Bellman equation at the heart of these algorithms \citep{roijers2013survey}, and therefore it may be necessary to condition the Q-values and the agent's choice of action on an \emph{augmented state} formed by concatenating the environmental state with the summed rewards previously received by the agent \citep{geibel2006reinforcement}. Additionally these approaches may fail to converge to the optimal policy in environments with stochastic state transitions \citep{vamplew2021stochasticity}.

An alternative to these value-based approaches is to adopt a policy-search algorithm. These have the advantage that by optimising at policy level, they can directly optimise with regards to any utility function, including non-linear functions. In addition, they generally produce stochastic policies, which can be beneficial in the context of multiple objectives as discussed earlier in Section \ref{sec:det-vs-stoch-policies}. For example, \citet{pan2020additional} implement a mixture of long-term policy gradient and short-term planning to find single-policy solutions, while \citet{siddiquelearning} develop multi-objective forms of the PPO and A2C policy search methods for the task of finding a single-policy which is fair with regards to all objectives, as measured by the Generalized Gini social welfare function. A substantial number of further multi-objective policy-search methods have been explored in the literature, but much of this work has been in the context of multi-policy approaches and/or deep RL, and so will be discussed further in the later sub-sections.

Under the ESR criterion (Section \ref{sec:seresr}) a non-linear utility function is assumed. As already highlighted a non-linear utility function invalidates the assumed additive returns in the Bellman equation. In this case, new methods must be created to efficiently optimise the ESR criterion.  \citet{roijers2018esr} implement an Expected Utility Policy Gradient (EUPG) algorithm which uses Monte Carlo simulations to calculate the sum of the accrued returns and future returns. EUPG optimises over the full returns of an episode as the utility function is applied to the sum of the accrued returns and the future returns. \citet{hayes2021dmcts,hayes2021dmcts_long} propose an algorithm known as Distributional Monte Carlo Tree Search (DMCTS) which learns a posterior distribution over the utility of the returns of a full episode and achieves state-of-the-art performance under the ESR criterion.

\subsubsection{Multi-policy approaches}
\label{sec:multi-pol}

Multi-policy approaches can be divided into two classes. \emph{Outer loop} methods operate on series of single-objective problems, whereas \emph{inner loop} methods consist of algorithms directly designed to produce multiple policies \citep{roijers2017multi}.

The simplest outer loop methods iterate through a series of different parameter settings for a utility function, and re-run a single-policy MORL method for each setting (for example, \citep{parisi2014policy}). The efficiency of outer loop approaches can be improved in two ways. Re-using information from earlier runs rather than discarding this information can reduce learning time \citep{natarajan2005dynamic,parisi2017manifold}. Secondly, naive searches through parameter space may re-learn the same policy multiple times, or require a small step-size to ensure all optimal policies are discovered \citep{roijers2015point}.  More efficient adaptive search methods can reduce the number of iterations of the outer loop \citep{roijers15computing,roijers2015point,roijersPhD}. 

Inner loop methods modify the underlying algorithm to directly identify and store multiple-policies in parallel rather than sequentially. Both Pareto-Q-Learning (PQL) \citep{van2014multi} and PQ-learning \citep{ruiz2017temporal} modify Q-learning to store multiple Pareto-optimal values for each state-action pair. Pruning of dominated values is used to eliminate dominated policies \citep{mandow2018pruning}. So far these methods are restricted to tabular representation of Q-values, limiting their broader applicability, although the Pareto DQN algorithm \citep{reymond2019pareto} provides an initial attempt to integrate PQL and deep RL methods. In the batch setting, Multi-Objective Fitted Q-Iteration (MOFQI)~\citep{castelletti2012tree} extends the Fitted Q-Iteration algorithm~\citep{ernst2005tree} to the multi-objective case by adding to the state the linear scalarisation weights. MOFQI learns with a single training process an approximation of the optimal Q-function for all the combinations of the scalarisation weights. 

Multiple authors have developed inner-loop multi-policy methods based on multi-objective extensions of Monte Carlo Tree Search. The decision about which branch of the tree to expand at any point is determined based on either the hypervolume metric, or on a measure based on Pareto-dominance \citep{wang2012multi,wang2013hypervolume,perez2013online,chen2019pareto,weng2020pareto}. 

Model-based methods have clear benefits in the context of multi-policy learning, as once a model of the environment has been learned, it can be used to derive the optimal policy for any utility function with no requirement for further interaction with the environment. Despite this, there has been surprisingly little research so far in model-based MORL. \citet{wiering2014model} provide an approach which learns all Pareto-optimal policies by first learning a model, and then applying the CON-MDP multi-objective dynamic programming algorithm \citep{wiering2007computing}. However, this approach is limited to learning stationary, deterministic policies for deterministic environments. The approach of \citet{yamaguchi2019model} can be applied to stochastic environments. It learns a model which stores reward occurrence probability (ROP) vectors rather than Q-values, and then uses the inner product of the ROP vector and a given weight vector to find the expected reward for the optimal policy for that weight vector. This approach avoids the need to perform an extensive search of the weight space to identify optimal policies. However, it is limited to finding deterministic policies under linear scalarisation, and is designed for maximising the average reward rather than the cumulative discounted return.

In order to learn in domains with continuous state-action spaces and where the state is not fully observable, policy search or actor-critic algorithms are usually considered~\citep{deisenroth2013survey}. In the literature, both outer loop~\citep{parisi2014policy} and inner loop~\citep{parisi2016multi,giuliani2016curses,parisi2017manifold} approaches have been proposed to extend policy search methods to multi-objective problems. The approach of \citet{parisi2017manifold} is interesting in that it constructs a continuous rather than discrete approximation of the Pareto front.

Population-based evolutionary methods are well-suited to finding multiple policies, as each individual can represent a policy which is optimal for a different set of utility preferences. The field of multi-objective evolutionary optimisation is already very well established \citep{trivedi2016survey,antonio2017coevolutionary,falcon2020indicator}, and several researchers have applied concepts from this area to MORL tasks. Evolutionary methods can either be applied directly \citep{cheng1988algorithms,parisi2017manifold}, or combined in a hybrid algorithm with local hill-climbing \citep{soh2011evolving}, policy-gradient \citep{xu2020prediction} or actor-critic methods \citep{chencombining}.

\subsubsection{Interactive approaches}

The majority of MORL methods take either an \emph{a priori} approach to policy selection where user's preferences must be specified prior to learning, or an \emph{a posteriori} approach where a set of policies are learned and then presented to the user for selection. A third alternative is to allow the user to interactively specify their preferences during the learning process, as first proposed in \cite[p.~63]{vamplew2011empirical}. This allows the user to make a more informed decision based on the agent's discoveries about the nature of achievable trade-offs between objectives, while also allowing earlier convergence to the user's preferences which is important in online learning. An example is the Q-steering algorithm \citep{vamplew2015reinforcement,vamplew2017steering}. The user specifies initial preferences in terms of a target point in objective space, and the agent learns a non-stationary mixture of linear-scalarised base policies which minimises the distance between the average reward and the target. A visualisation of the returns of the base policies can be provided to the user, who may then revise their choice of target. The agent can immediately adapt to such changes.

Some work builds on methods for single-objective reinforcement learning with human guidance, and extends those methods to multi-objective problems. \citet{wanigasekara2019learning} propose an algorithm that learns user utility functions from observations of user-system interactions for multi-objective contextual bandit based personalized ranking of search results. \citet{ikenaga2018inverse} propose to use inverse reinforcement learning for elicitation of user preferences in multi-objective sequential decision making, while \citet{saisubramanian2020multi} use human feedback through random queries, approval, corrections, and demonstrations to learn policies that avoid negative side effects.

A systematic approach to simultaneous learning about the environment and the user was proposed by \citet{roijers2017interactive} for multi-objective multi-armed bandits. Specifically,  the \emph{interactive Thompson sampling (ITS)} algorithm  uses queries to solicit preferences resulting from linear utility functions while interacting with the environment. For this, it employs Bayesian logistic regression to learn about the utility function, and uses the uncertainty estimates about the utility function to decide which queries to ask. The \emph{Gaussian-process Utility Thompson Sampling (GUTS)} algorithm \citep{roijers2020interactive} does the same for any continuous utility function by using Gaussian processes to model the utility function, and estimate the uncertainty about this function.

\subsubsection{Scaling up to high-dimensional states}

Single-policy and outer loop multi-policy methods can be extended to handle high-dimensional input data in much the same way as the corresponding single-objective algorithms on which they are based. For example, \citet{tesauro2008managing} combined SARSA, non-linear utility, and small multilayer perceptrons to learn to control the power consumption and performance of computing clusters. 

Deep reinforcement learning methods~\citep{mnih2015human,lillicrap2015continuous} have shown to scale beyond finite and discrete spaces to problem domains with high-dimensional, continuous state and action spaces. Using deep networks as non-linear function approximators for handling multi-objective optimization problems has been on the rise the past few years~\citep{mossalam2016multi,li2020deep,abels2019dynamic,tajmajer2018modular,nguyen2020multi}. Most of these methods extend the single-objective DQN architecture~\citep{mnih2015human} and follow a single-policy or a multi-policy approach.

\citet{mossalam2016multi} extended DQN to a multi-objective setting by learning an approximate coverage set of policies (multi-policy). Each policy is represented using a DQN whose output layer has $\mathcal{|A|} \times n$ nodes, where $\mathcal{|A|}$ represents the size of the action space and $n$ represents the number of objectives. For better efficiency, the authors proposed to re-use the network weights of previously learnt policies for preferences that are similar to each other ($\boldsymbol{w'} \sim \boldsymbol{w}$). \citet{abels2019dynamic} analyzed different architectures while extending DQN to a multi-objective setting: a multi-policy approach with multiple DQNs for different user preferences; and a single-policy approach with a single DQN that can generalize across different user preferences. In scenarios where user preferences change in real-time, the single-policy method was most effective. To improve sample efficiency and address bias to recently seen user preferences, they used a diverse experience replay buffer that contained experiences corresponding to different user preferences. \citet{yang2019generalized} also used a single-policy approach which generalized across different user preferences, however, they performed envelop updates by using a convex envelope of the solution frontier while updating network parameters. Such envelop updates lead to faster convergence when compared to scalarized updates for a given user preference, which are often sample inefficient, resulting in sub-optimal policies. 

There has been some recent work on multi-objective deep reinforcement learning for continuous action spaces.
\citet{chen2019meta} combined a multi-objective extension of PPO \citep{schulman2017proximal} with model-agnostic meta-learning (MAML) \citep{finn2017model}. The proposed method first learns a meta-policy, which can then be fine-tuned in few iterations to find a set of Pareto optimal policies. Compared to learning each policy from scratch, the method was shown to improve performance in terms of training time and optimality of the resulting Pareto front. \citet{xu2020prediction} also used a multi-objective extension of PPO, combined with an evolutionary algorithm to guide learning in the most promising direction. In each generation of evolution, data stored from previous iterations of MORL are used to fit a prediction model, which can help find the pairs of policies and scalarization weights that will improve the solution the most. Each selected policy-weight pair is then improved through MORL to produce offspring policies, which are used to create a new generation of policies. The final generation is divided into policy families by clustering, and policy parameters within each family are interpolated to produce a continuous approximation of the Pareto front. \citet{abdelfattah2019intrinsically} used a two-stage approach for learning in environments with non-stationary dynamics and continuous actions. In the first stage, a set of generic skills (e.g., \textit{Move Forward}, \textit{Turn Left}, and \textit{Turn Around}) are learned. In the second stage, the learned skills are used in a hierarchical version of DDPG \citep{lillicrap2015continuous} to produce a policy coverage set for the MOMDP. An intrinsically motivated RL algorithm is used to select which objective preferences to explore to improve the coverage set, and a policy bootstrapping mechanism is used to quickly adapt to changes in the environment dynamics.

While the above approaches use linear scalarization, Tajmajer \citep{tajmajer2018modular} proposed a non-linear action-selection mechanism by using $n$ different DQNs corresponding to each objective which are combined using a separate output layer along with the user preferences. Deep reinforcement learning methods for multi-objective partially observable settings have also been proposed~\citep{nian2020dcrac}. These approaches use action and observation histories along with user preferences as input to the neural network. In general, partially observable settings are much more complex when compared to fully observable settings in terms of training time as well as training stability.

\subsubsection{Multi-agent algorithms}
\label{sec:momas_algos}

\begin{figure}[ht!]
    \centering
    \includegraphics[width=0.9\textwidth]{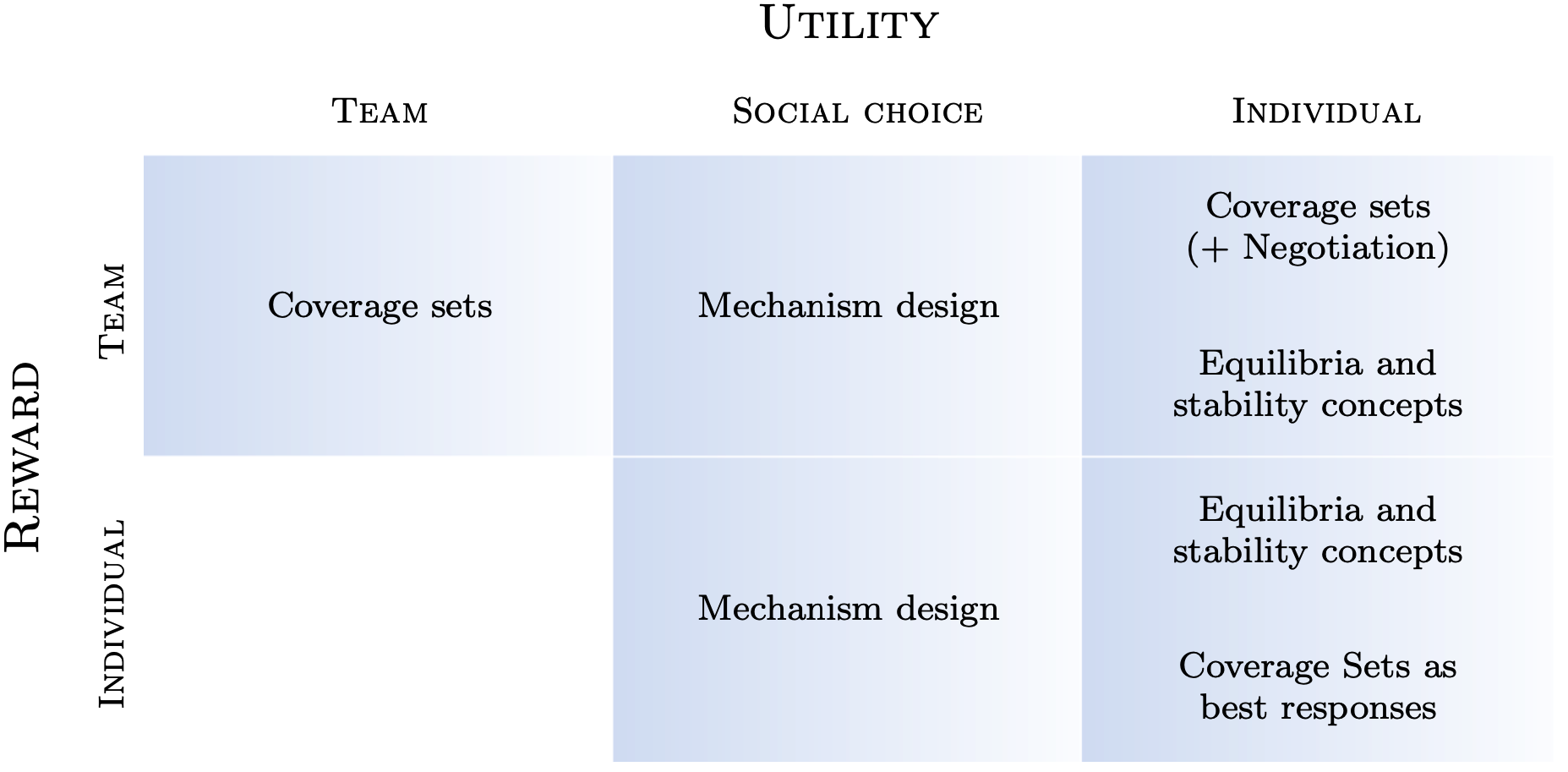}
    \caption{Multi-objective multi-agent decision making taxonomy and mapping of solution concepts \citep{radulescu2020survey}.}
    \label{fig:momarl_tax}
\end{figure}

As explained in Section~\ref{sec:utility_based}, in single-agent multi-objective problems, the shape of the utility function, in conjunction with the allowed policy space, can be used to derive the optimal solution set that a multi-objective decision-theoretic algorithm should produce. In multi-agent settings, the situation is more complex, as each individual agent can represent one or more distinct users (i.e., each agent can have a different utility function). For this reason, \citet{radulescu2020survey} proposed a new taxonomy which classifies multi-objective multi-agent decision making (MOMADM) settings on the basis of both reward structures and utility functions, as shown in Figure~\ref{fig:momarl_tax}. We note that the case of individual reward--team utility is equivalent to and treated as the individual reward--individual utility case, since the individual return vectors would still lead to different utility values for each agent, despite them having the same utility functions.

In multi-objective multi-agent settings the agents' strategies are interrelated. For this reason, \emph{solution concepts}, i.e., whether the agents in the system reach outcomes that are of interest, could be used to evaluate the algorithms' performance. We detail below the solution concepts identified by \citet{radulescu2020survey} for the MOMADM setting and present a few algorithmic approaches that employ them.

\emph{Coverage sets}. The team reward and team utility setting in MOMADM represents a fully cooperative scenario, where all agents share the same rewards and derived utility. Since there is only one true utility function in the execution phase, coverage sets represent the right solution concept for this case, with the same motivation as for single-agent multi-objective decision making. Multi-objective coordination graphs (MOCoGs) represent one of the most studied models for cooperative multi-objective multi-agent systems. They exploit the fact that in multi-agent systems the rewards or agents can often be factorised into smaller components. Numerous algorithmic approaches focus on finding (approximate) Pareto coverage sets (\ref{sec:solsets}) like for example multi-objective bucket elimination (MOBE) \citep{rollonThesis,Rollon06MOBE}, multi-objective Russian doll search \citep{rollon2007multi}, multi-objective (AND/OR) branch-and-bound tree search \citep{marinescu2009exploiting,marinescu2011efficient,rollon2008constraint}, Pareto local search \citep{inja2014queued}, and multi-objective max-sum \citep{delle2011bounded}. Another frequently used model is the cooperative multi-objective stochastic game (MOSGs), where reinforcement learning or evolutionary algorithms were used to derive coverage sets (e.g., \citep{mannion2018reward,Mannion2017Policy,Yliniemi2016Multi}). Similar methods were proposed for the individual reward and utility setting, where a coverage set can also be a set of possible best responses to the behaviours of the other agents (e.g., \citep{avigad2011optimal,eisenstadt2015co,dusparic2009distributed}). In an individual reward--team utility setting, coverage sets could be used if all agents agree (e.g., through negotiation \citep{jonker2017automated}) upon which alternative joint policy from the coverage set to execute. 

\emph{Equilibria and stability concepts}. In the individual utility scenario, the utility derived by each agent from the received reward is different, regardless if this reward is the same or not for all the agents. Suitable solution concepts for dealing with decision making between self-interested agents are game theoretic equilibria (e.g., Nash equilibria \citep{Nash1951Non}, correlated equilibria \citep{aumann1987correlated}). We find here works that study the idea of robust equilibria in multi-objective games \citep{qu2015robust,yu2013robust} or how equilibria are affected by the use of the different optimisation criteria \citep{radulescu2020utility}. Furthermore, knowledge transfer \citep{Taylor2014Accelerating} and opponent modelling \citep{zhang2020opponent} also becomes more important in this context.

When binding agreements among agents are possible, solution concepts from cooperative game theory can also apply to individual utility settings.  Coalition formation can therefore become a central problem in these cases, i.e., finding (sub)groups of agents that are willing to make such a binding agreement with each other \citep{igarashi2017multi}. 

\emph{Mechanism design}. In game theory, the field of mechanism design takes the system's perspective for multi-agent decision problems. This implies taking as input both the original decision problem (where the agents have individual reward functions that are unknown to the other agents and the ``owner'' of the game), as well as a social welfare function. The aim is to design a system of additional payments that would (a) force the agents to be truthful about their individual utilities, and (b) lead to solutions that are (approximately) optimal under the social welfare function. In multi-objective settings, the situation is more complex, as the individually received rewards determine the utilities via individual, private utility functions. In general, it can be very challenging?, or even impossible to articulate these functions, so being ``truthful'' about one's utility might be infeasible from the get-go.  Nevertheless, it is possible to design mechanisms for some multi-objective multi-agent problems if the individual utilities can be articulated (e.g., \citep{grandoni2010utilitarian,pla2012multi,Ramos+2020aamas}).

For an in-depth overview of solution concepts for multi-objective multi-agent decision making, the interested reader is referred to a recent survey by \citet{radulescu2020survey}.

\section{Evaluating the performance of multi-objective decision making algorithms}
\label{sec:evaluating_performance}

Unlike in single-objective RL, there is not only one optimal solution in multi-objective problem settings. 
MORL algorithms therefore often produce solution \emph{sets} (see Section \ref{sec:solsets}).
This complicates the evaluation and comparison procedure of MORL algorithms: 
When is one solution set better than the other? 
What properties should a solution set have, and how do we measure those?

In this section, we give an overview of existing evaluation metrics, 
starting with \emph{axiomatic-based} ones (Section \ref{sec:evaluation:axiomatic}).
These assume that the optimal solution is the true Pareto front (or convex hull), and try to compare to this in aspects like spread, coverage, or distance.
However, these axiomatic metrics are often difficult to interpret from a user perspective.
As argued in Section \ref{sec:utility_based}, the development of MORL solutions should be driven by the perspective on user utility.
Similarly, \emph{utility-based} evaluation metrics should be used when assessing MORL algorithms (Section \ref{sec:eval_utility}).

After giving an overview on evaluation metrics and approaches, we briefly discuss discuss potential pitfalls when using value function approximations in MORL settings in Section \ref{sec:evaluation:caution}.
Section \ref{sec:benchmarks} gives an overview of existing benchmarks and their properties.

\subsection{Axiomatic-based evaluation metrics}
\label{sec:evaluation:axiomatic}
In this section we give an overview of axiomatic approaches to evaluating solutions to multi-objective decision making problems. Such approaches were widely-used in early literature in the field.

\subsubsection{The hypervolume metric}
\label{sec:metric-hypervolume}

\begin{figure}[ht]
    \centering
     \begin{subfigure}[b]{0.4\textwidth}
         \centering
         \includegraphics[width=\textwidth]{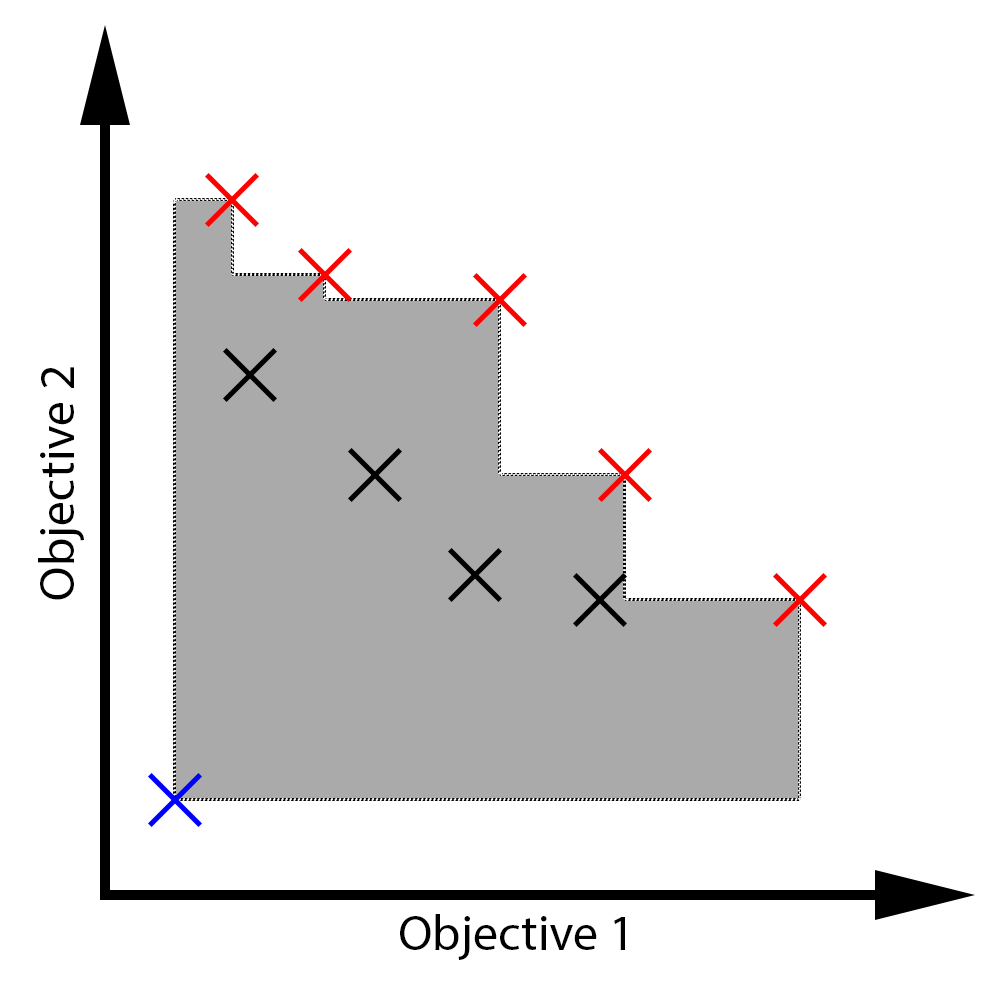}
         \label{fig:hypervolume_original}
     \end{subfigure}
     \hfill
     \begin{subfigure}[b]{0.4\textwidth}
         \centering
         \includegraphics[width=\textwidth]{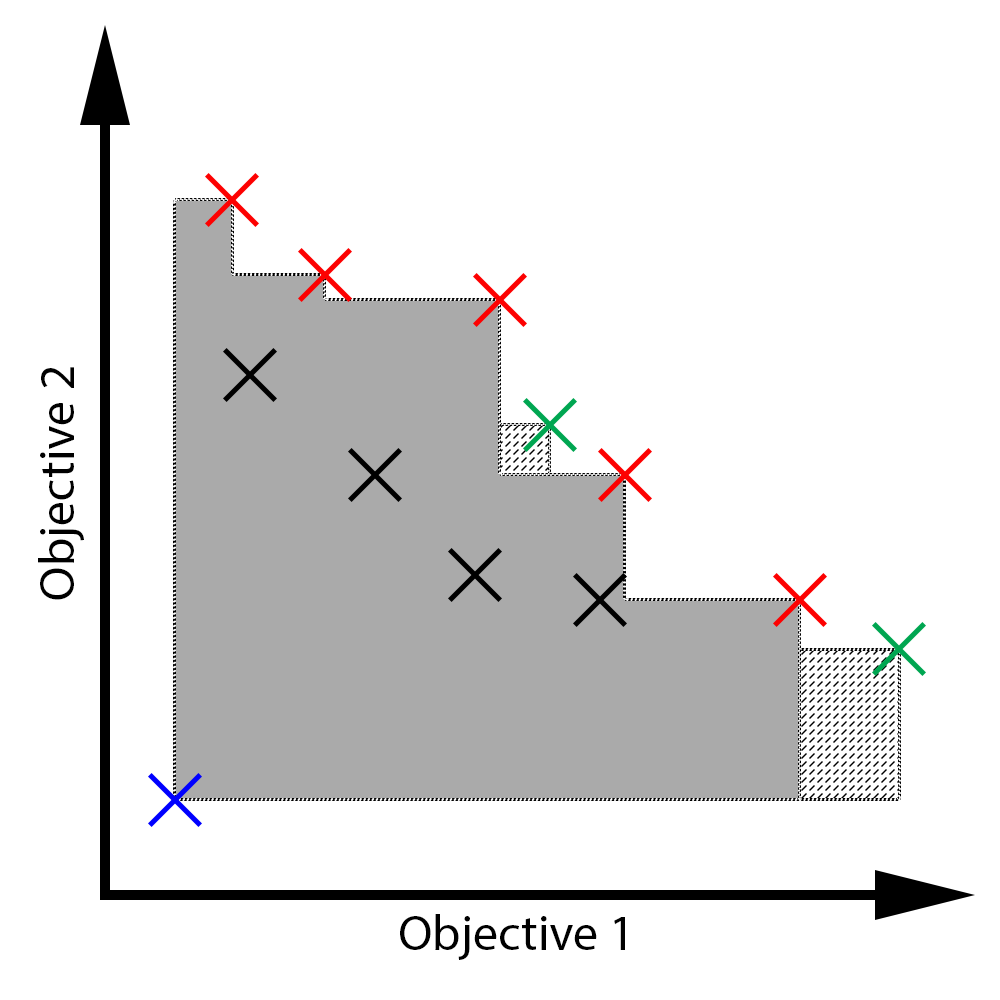}
         \label{fig:hypervolume_extra_points}
     \end{subfigure}
     \hfill
    \caption{Left: A graphical illustration of the hypervolume for a 2-objective problem, where both objectives are to be maximised. Solutions in red form the undominated set, while solutions in black are said to be dominated. The shaded area denotes the hypervolume of the undominated set with respect to the reference point (shown in blue). Right: The effect of adding two new points (shown in green) to the undominated set.}
    \label{fig:hypervolume}
\end{figure}

The hypervolume metric has been widely used to evaluate the performance of multi-objective decision making algorithms (e.g., \citep{mannion2018reward,vamplew2011empirical,van2014novel,van2013hypervolume,wang2013hypervolume,Yliniemi2016Multi}).
The hypervolume metric measures the (hyper-)volume in value-space Pareto-dominated by the set of policies in an approximate coverage set. This correlates with (but is not equal to) the spread of a set of undominated solutions over the possible multi-objective solution space. For this reason, it has been used to compare the sets of solutions produced by multi-policy algorithms (or indeed single policy algorithms run multiple times with different scalarisation/utility function parameters). 
The accuracy of any set of solutions produced by an algorithm can be evaluated by comparing its hypervolume with that of the non-dominated set produced by a competing algorithm, or with that of the true Pareto front of the application domain (if known). In domains where the true Pareto front is known, the hypervolume represents an absolute maximum level of performance that may be achieved in terms of coverage of the set of solutions over the objective space:
\begin{equation}\label{eq:hypervolume}
 \mathtt{HyperVolume}(CS,{\bf V}_{\text{ref}}) = \bigcup_{\pi \in CS} \mathtt{Volume}({\bf V}_{\text{ref}}, {\bf V}^\pi),
\end{equation}
where $\mathtt{Volume}({\bf V}_{\text{ref}}, {\bf V}^\pi)$ is the volume of the hypercube spanned by the reference vector, ${\bf V}_{\text{ref}}$, and the vector in the CS, ${\bf V}^\pi$.

Figure~\ref{fig:hypervolume} illustrates the hypervolume of a set of undominated solutions with respect to a given reference point, ${\bf V}_{\text{ref}}$, for a 2-objective maximisation problem, where both objectives are to be maximised. For convenience, a reference point in the multi-objective space is often used when calculating the hypervolume of a non-dominated set. This reference point may be chosen arbitrarily. 

Although widely used in the literature, the hypervolume metric has a number of problems. The most significant of these is that hypervolume values are difficult to interpret, as they do not map to any real-world notion of value or utility. When comparing the hypervolume of two competing sets of solutions, the benefit of a certain increase or decrease in hypervolume is not readily apparent to the end user. Adding just one non-dominated solution at the extreme ends of the objective ranges could lead to a large increase in the hypervolume of a non-dominated set, even if this additional solution is of little interest to the end user. Conversely, adding a new solution that is close to other solutions in the non-dominated set can result in a minimal increase in hypervolume, even if the new solution is valuable to the end user. Finally, it is unlikely that the true set of non-dominated solutions will be known a priori for any non-trivial multi-objective decision making applications. This invalidates one of the main arguments for the use of the hypervolume metric, i.e., evaluating the coverage of a set of solutions with respect to a reference set. Furthermore, the hypervolume is only applicable to settings where every Pareto-non-dominated policy potentially contributes to the utility. This is not always the case. For example, when the utility function is known to be linear, the hypervolume is not applicable as many policies that would contribute to the hypervolume are known to not improve utility (i.e., all concave regions in the Pareto front). For these reasons, we recommend the use of alternative metrics that better reflect the usefulness of the solutions produced by an algorithm (such as the user's utility).

\subsubsection{Sparsity of coverage sets}
The information that is contained by metrics like the hypervolume is rather limited. The only guarantee we have is that if the hypervolume is maximised (unless there are points that contribute 0 hypervolume at the edges that we have missed), then a Pareto Coverage Set has been recovered. This is of course not informative, especially during learning. 

One key bit of critique is that if we have two approximate solution sets with equal -- or approximate -- hypervolume, then we should prefer the set which has more spread over the value space. In other words, the set that contains value vectors that are furthest apart from each other is the better one. From a utility-based perspective, this is also intuitive, as the user will pick the best vector from a solution set $\mathcal{S}$ according to:
\begin{equation}
\pi^* = \arg\max_{\pi \in \mathcal{S}} u({\bf V}^\pi),
\end{equation}
so it helps if the user has a larger variety of value vectors to select from.\footnote{Please note that this intuition implicitly assumes some form of continuity in the user's utility function, $u$.}

In multi-objective optimisation, this idea has been used to create algorithms that explicitly look for diverse solutions \citep{deb2002fast}. In MORL, this same idea has been used to diversify the experience replay buffer, in order to be able to adapt to different utility functions faster \citep{abels2019dynamic}.  

In addition to finding solution sets that are evenly spread over the value space (i.e., sets with high diversity), it is desirable that the solutions provide a dense coverage of the whole Pareto front (i.e., sets with high resolution). For this purpose, \citet{xu2020prediction} proposed to use sparsity as a metric for evaluation of Pareto front approximations. The proposed sparsity metric is defined as:

\begin{equation}
    Sp(\mathcal{S}) = \frac{1}{|\mathcal{S}| - 1}\displaystyle\sum_{j=1}^{m}\displaystyle\sum_{i=1}^{|\mathcal{S}|-1}(\tilde{\mathcal{S}}_j(i)-\tilde{\mathcal{S}}_j(i+1))^2, 
\end{equation}

Here $\mathcal{S}$ is the Pareto front approximation for an environment with $m$ objectives, and $\tilde{\mathcal{S}}_j(i)$ is the $i$-th value in the sorted list for the $j$-th objective values in $\mathcal{S}$. According to this metric, Pareto front approximations with lower sparsity are better.

\subsubsection{The $\varepsilon-$metric}

The $\varepsilon$ metric \citep{zitzler2008quality} measures how closely a solution set $\mathcal{S}$ approximates the Pareto front $PF$. 
It has been widely used in multi-objective evolutionary optimisation \citep{zitzler2008quality} and reinforcement learning \citep{vamplew2017softmax}.
There are two measures, the additive and the multiplicative $\varepsilon$-indicator. 

The \textbf{additive} $\varepsilon$-indicator is given by
\begin{equation}
    I_{\varepsilon^+} = 
    \inf_{\varepsilon\in\mathbb{R}} 
    \{ 
    \forall \mathbf{V}^\pi\!{\in}\ PF, ~
    \exists \mathbf{V}^{\pi'}\!{\in}\ \mathcal{S} :
    V_i^\pi \le V_i^{\pi'} + \varepsilon, \forall i \in \{1,\dotsc,n\} 
    \}, 
\end{equation}
where $n$ is the number of objectives.
where $\mathbf{V}^\pi\in\mathbb{R}^d$ is the $d$-dimensional value of policy $\pi$ (see Section \ref{sec:bg}).
In words, a solution set $\mathcal{S}$ is an $\varepsilon$-approximate Pareto front according to this metric if \emph{for each} value vector $\mathbf{V}^\pi$ on the Pareto front $PF$, there \emph{exists at least one} value vector $\mathbf{V}^{\pi'}$ in the solution set $\mathcal{S}$, such that for each objective $d$, the value in $\mathbf{V}^{\pi'}$ is \emph{at most} $\varepsilon$ smaller than the values in $\mathbf{V}^\pi$.

The (less commonly used) \textbf{multiplicative} $\varepsilon$-indicator is given by
\begin{equation}
    I_{\varepsilon^*} = 
    \inf_{\varepsilon\in\mathbb{R}} 
    \{ 
    \forall \mathbf{V}^\pi\!{\in}\ PF,  ~
    \exists \mathbf{V}^{\pi'}\!{\in}\ \mathcal{S} : 
    V_i^\pi \le V_i^{\pi'} (1 + \varepsilon), \forall i \in \{1,\dotsc,n
    \}\} ,
\end{equation}
The difference to the additive indicator is in how the distance is calculated: here, each objective can at most be worse by a multiplicative factor of $1+\varepsilon$, i.e., this scales with the magnitudes of the individual values (objectives with larger values allow a larger deviation).

The $\varepsilon$ metric gives an indication of the factor by which an approximate solution set is worse than the Pareto Front, considering all objectives.
It can also be used to prepare two arbitrary solution sets instead of a solution set and the Pareto front.
Unlike the hypervolume metric, it can give an indication of \emph{whether} one is better than the other (they might, however, be incomparable w.r.t. this metric).

We argue that the $\varepsilon$ metric is more useful than the hypervolume, since it can directly be used to derive a utility for a given user, see Section \ref{sec:eval_utility}.

\subsubsection{Metrics from information retrieval}

The Coverage Ratio metric is used in \citep{yang2019generalized} as an evaluation metric for comparing different multi-objective algorithms. It is a measure of the count of policies recovered from a finite Coverage Set (CS) which is determined by a comparison between the set $\mathcal{S}$ of policies $\phi$ with value vectors $\mathbf{V}^{\pi}\in\mathbb{R}^d$ found by a MORL algorithm, and the value vectors corresponding to the policies in the (ground-truth) CS. The measure weights both the precision and recall of finding policies in the CS, the following definition is used when calculating precision and recall such that policies with value vectors within epsilon of the value vector of a policy in CS, are classed as in the CS.

\begin{equation}
    \mathcal{S}\cap_{\epsilon} CS = 
    \{ 
    \mathbf{V}^\pi \in \mathcal{S}\mid ~
    \exists \mathbf{V}^{\pi^*} \in CS ~
    \text{:}~
    \lVert \mathbf{V}^\pi-\mathbf{V}^{\pi^*} \rVert_{1}/\lVert \mathbf{V}^{\pi^*} \rVert_{1}    < \epsilon
    \}
\end{equation}
The Coverage Ratio (also known as the F-score) is then calculated as the harmonic mean between the $\text{precision} =|\mathcal{S}\cap_{\epsilon} CS|/|\mathcal{S}|$  and $\text{recall} =|\mathcal{S}\cap_{\epsilon} CS|/|CS|$ measures.
\begin{equation}
\label{eq:CR}
    CR(\mathcal{S}) = 
    2 \cdot\frac{\text{precision} \cdot \text{recall}}{\text{precision}+\text{recall}}
\end{equation}

We argue that there are several issues with the Coverage Ratio metric. Firstly, the measure (Equation \ref{eq:CR}) implies that precision and recall are equally important which in reality is not the case. For example, if the utility function is linear, and $\mathcal{S}$ contains excess policies that have a value which is the weighted sum of two other value vectors. Such vectors need not be in the CCS (convex coverage set, see Equation~\ref{eq:ccs}) and decreases the precision. However, having this excess policy does not decrease the utility for any linear utility function. Conversely, as missing out a policy in the CS typically does decrease the utility for a whole range of utility functions.

Secondly, like the hypervolume metric, the Coverage Ratio does not account for the different levels of utility a user gains from different solutions. In this measure the presence of any solution from the CS is treated as of equal value to any other solution found also in the CS. However, unlike the hypervolume metric the Coverage Ratio is not correlated at all with the spread of the set of non-dominated solutions over the possible multi-objective space. Therefore, this measure fails to account for any utility the user gains from the spread of solutions but retains the undesirable properties of the hypervolume metric. 

Thirdly, the $\varepsilon$-parameter that controls the threshold for when we consider a policy in the Convex Coverage Set is a parameter that needs to be chosen and can have a large impact on the Coverage Ratio. Specifically, if a value estimate $\tilde{\bf V}$ is within the hypercube surrounding a value in the CS, ${\bf V}_i \pm \varepsilon$, it is assumed to correspond to that value vectors in the CS. Setting $\varepsilon$ arbitrarily high can lead to all solutions being treated in the CS, while setting it low could lead to an algorithm producing no solutions in the CS. When comparing different algorithms, the choice of $\varepsilon$ could have a large impact on the final ranking and it is not a priori clear what a fair setting of $\varepsilon$ would be.

Lastly introducing the notion of $\varepsilon$ to the recall measure means that multiple counting of policies in the CS can occur as more than one value vector in ${F}$ could be within $\varepsilon$ distance of the same policy's value vector in the CS. Not only can an algorithm thus ``recall'' more than the ground truth number of policies, but more importantly, ``recalling'' two policies for the same ground truth policy can obfuscate the missing of another policy in the resulting value of the metric, which is of course highly undesirable.

\subsection{Utility-based evaluation metrics} \label{sec:eval_utility}

As argued in Section~\ref{sec:utility_based}, the utility-based approach is preferable in most scenarios, since here the algorithms are designed and evaluated with respect to the utility that the solution can offer to the user. 
Accordingly, any evaluation metric should take this into account. 
Many of the axiomatic-based evaluation metrics are difficult to interpret in terms of user utility, and in addition they often require access to the true Pareto front. 
If it is possible to assess the utility of the user at time of deployment, then solution sets can be compared based on user utility. 

For instance, the user's utility might correspond to revenue that a deployed solution achieves; in this case, the utility can be measured and compared directly.

For when this is not possible, \citet{zintgraf2015quality} propose two utility-based evaluation methods, the \emph{expected utility metric} (EUM) and \emph{maximal utility loss} (MUL). Compared to many other metrics such as the hypervolume metric, these are more suitable to compare different algorithms, since they are aimed at directly evaluating an agent's ability to maximize user utility, which is always our ultimate goal.

For a given solution set, the EUM is defined as the expected utility for a user from this solution set, under some prior distribution over user utility functions. 
Under the SER optimality criterion, this can be written as:
\begin{equation}\label{eq:eum}
    \text{EUM} = \mathbb{E}_{P_u}\left[\max_{\pi \in \mathcal{S}} u({\bf V}^{\pi})\right],
\end{equation}
where $\mathcal{S}$ the solution set outputted by an algorithm, 
$u$ the utility function of the user,
${\bf V}^\pi$ the vector-value of the best policy from that set (according to $u$).
The expectation is taken with respect to the distribution over utility functions $P_u$.
This metric is useful in situations where we care about the agent's ability to do well across many different utility functions, e.g., because many policies from the solution set will be used over time, or because they will be used for different users. This metric does however require a good prior over possible scalarisation functions in order to meaningfully evaluate a given solution set.

The MUL measures the maximal loss in utility that occurs when taking a policy from a given solution set, instead of the full set of possibly optimal solutions.
Under the SER optimality criterion, this can be written as:
\begin{equation}
    \text{MUL} = \max_{u\in\mathcal{U}} \left(\max_{\pi^* \in \mathcal{S}^*} u({\bf V}^{\pi^*}) - \max_{\pi \in \mathcal{S}} u({\bf V}^\pi)\right), 
\end{equation}
where $\mathcal{S}^*$ is the true optimal solution set (PF or CH, or a very good approximation thereof), $\mathcal{S}$ the solution set outputted by an algorithm, ${\bf V}^\pi$ the vector-value of said policy, and $u$ the utility function of the user, over which we take the maximisation with respect to the space of possible utility functions $\mathcal{U}$.  
Since it is often infeasible to compute the full set of optimal solutions in order to compute this metric and compare algorithms, a good reference set can be used (such as the union of multiple solution sets, e.g., the final solution sets of all algorithms evaluated in a comparison). 

We note that MUL is bounded if an $\varepsilon$-bound can be given on the accuracy of the set $\mathcal{S}$ produced by the algorithms, and the utility function is guaranteed to be (Lipschitz)-continuous \citep{zintgraf2015quality}.

\subsection{A word of caution regarding value vector approximation}
\label{sec:evaluation:caution}

In multi-objective planning, it is often the case that for a given policy, $\pi$, we know the exact value vector, ${\bf V}^\pi$. When the human decision maker (or even another algorithm) selects such a policiy in the selection phase (see Section \ref{sec:design_factors:motivating_scenarios}), we can thus trust these value vectors. This is key as we derive utility from these value vectors by applying the utility function to them (either implicitly or explicitly).

In multi-objective reinforcement learning it is tempting to think we have proper value vectors too, as many algorithms produce value vector estimates. In the literature these are often also denoted ${\bf V}$ or ${\bf Q}$. However, it is essential to note that these are stochastic estimates, that may well have both high variance, or even systematic biases \citep{hasselt2010double}. This issue is exacerbated by the use of function approximators, such as neural networks, which may have their own added variance and/or biases. It would therefore be fairer to explicitly denote such value estimates as estimates by using tildes, $\tilde{\bf V}$ or $\tilde{\bf Q}$, for example, but this is not common practice. 

In multi-objective RL, having inexact value estimates in the coverage set presented to human decision makers (or other algorithms), can lead to missing on two sides: firstly, if the value estimate of the actual best policy is off, that policy may not be selected. Furthermore, the value estimate of the policy that is selected may also be off, leading to a different utility than expected. Combined, these two sources of potential loss can severely affect the user utility. 

In order to mitigate the issue of inexact value estimates, and maybe even more importantly, biases in value estimates, we recommend that the coverage sets presented in selection phases do not directly rely on the value vector estimates from the MORL algorithms. Instead, we recommend to extract the policies that constitute the coverage set, and run a separate and thorough policy evaluation, before selecting any policy to execute.

\subsection{Benchmark problems for multi-objective decision making} \label{sec:benchmarks}

Well established benchmark problems are important for evaluation of reinforcement learning algorithms, since even small variations in the experiment design may have a significant impact on the results. By using common benchmarks a fair comparison of different approaches can be ensured, and by evaluating algorithms on several benchmarks the generality of results can be studied. Table~\ref{table:benchmarks} presents an overview of frequently used MORL benchmarks with discrete states and actions, as well as more recent extensions and additions with high-dimensional states, partial observability, multiple agents, and continuous actions.

\begin{table}[t]
\caption{Benchmarks for multi-objective reinforcement learning.}
\label{table:benchmarks}
\begin{threeparttable}
\begin{tabular}{ |c|c|c|c|c|c|  }
 \hline
 \multicolumn{6}{|c|}{Benchmarks} \\
 \hline
 Name & \makecell{Number of \\Objectives} & \makecell{Observation \\Space} & \makecell{Action \\Space} & \makecell{Pareto \\Front} & Ref.\\
 \hline
 Deep Sea Treasure & 2 & Discrete & Discrete & Known & \citep{vamplew2011empirical}\\
 \hline
 \makecell{Deep Sea Treasure~\tnote{a}} & 2 & Continuous & Discrete & Known & \citep{mossalam2016multi}\\ 
 \hline
 \makecell{Deep Sea Treasure~\tnote{b}} & 2 & Continuous & Discrete & Known & \citep{nian2020dcrac}\\ 
 \hline
 \makecell{Deep Sea Treasure~\tnote{c}} & 2-3 & Continuous & Discrete & Known & \citep{hasan2019dynamic}\\ 
 \hline
 MO-Puddleworld & 2 & Discrete & Discrete & Known & \citep{vamplew2011empirical}\\
 \hline
 MO-Mountain-Car & 3 & Discrete & Discrete & Known & \citep{vamplew2011empirical}\\
  \hline
 Resource Gathering & 3 & Discrete & Discrete & Known & \citep{vamplew2011empirical}\\
  \hline
 Linked Rings & 2 & Discrete & Discrete & Known & \citep{vamplew2017steering}\\
  \hline
 Non Recurrent Rings & 2 & Discrete & Discrete & Known & \citep{vamplew2017steering}\\
  \hline
 Space Exploration & 2  & Discrete  & Discrete & Known & \citep{vamplew2017softmax}\\
  \hline
 Bonus World & 3 & Discrete & Discrete & Known & \citep{vamplew2017softmax}\\
  \hline
 \makecell{MO Beach Problem~\tnote{*}} & 2 & Discrete & Discrete & Known & \citep{mannion2018reward}\\
  \hline
 Mine Cart & $\geq$2 & Continuous & Discrete & Known~\tnote{d} & \citep{abels2019dynamic}\\
  \hline
 HalfCheetah-v2 & 2 & Continuous & Continuous & Unknown & \citep{xu2020prediction}\\
  \hline
 Hopper-v2 & 2 & Continuous & Continuous & Unknown & \citep{xu2020prediction}\\
  \hline
 Swimmer-v2 & 2 & Continuous & Continuous & Unknown & \citep{xu2020prediction}\\
  \hline
 Ant-v2 & 2 & Continuous & Continuous & Unknown & \citep{xu2020prediction}\\
  \hline
 Walker2d-v2 & 2 & Continuous & Continuous & Unknown & \citep{xu2020prediction}\\
  \hline
 Humanoid-v2 & 2 & Continuous & Continuous & Unknown & \citep{xu2020prediction}\\
  \hline
 Hopper-v3 & 3 & Continuous & Continuous & Unknown & \citep{xu2020prediction}\\
 \hline
\end{tabular}
\begin{tablenotes}
    \item[a] With image observations.
    \item[b] With image observations and partial observability.
    \item[c] With image observations and dynamic environment.
    \item[d] Optimal solutions are known for the default configuration of the environment.
    \item[*] Multi-agent environment.
\end{tablenotes}
\end{threeparttable}
\end{table}

For multi-objective decision problems other than MOMDPs, such as multi-objective coordination graphs \citep{marinescu2009exploiting,roijers2017multi,rollon2008constraint} and multi-objective normal form games \citep{radulescu2020utility,zhang2020opponent}, benchmarks are also few and spread out over different papers. 

\section{An illustrated example}
\label{sec:worked_examples}
In Section~\ref{sec:motivating_examples}, we illustrated diverse problems that require multi-objective optimisation. One of them is the water management problem (Section~\ref{sec:water}). In that case, as we want to propose a diverse set of solutions to the decision maker---following the taxonomy defined in Section~\ref{sec:design_factors}---we are in a multi-policy scenario. 
In this section, we concretely tackle the water management problem by applying a multi-objective algorithm, and comparing it with its single-objective counterpart, with both producing deterministic policies. We make no assumptions about the utility function of the decision maker, except that it is monotonically increasing. Thus, the solution set we aim to produce is the Pareto coverage set of deterministic non-stationary policies. 

\subsection{Setting}

In this problem, the goal is to control a dam responsible for dispatching water downstream while avoiding flooding in the region. This environment is modelled as a one-dimensional, continuous state, representing the amount of water present in the reservoir. This is subject to change, depending on factors such as rain. At each timestep, the dam can release a specified water amount (a one-dimensional, continuous action) \footnote{The code used to illustrate this setting can be found at the following link: \url{https://gitlab.ai.vub.ac.be/mreymond/morl-guide}}. 

The dam is responsible for supplying water, and needs to meet the water demand. At the same time, it should be careful not to hold too much water, as this increases the risk of flooding upstream. These objectives are conflicting, since the inflow of water is on average insufficient to cope with the water demand. In order to increase the chance of meeting future demand, the reservoir needs to be filled, thus increasing the risk of flooding upstream.

\subsection{Multi-objective natural evolution strategies}

With an unknown utility function, which may be non-linear, we want to propose a set of alternatives to the decision makers by approximating the Pareto-front. Since we have no prior information about the preferences of the decision makers, we make no assumptions about the utility function.

The algorithm that is used to approximate the coverage set is Multi-Objective Natural Evolution Strategies (MONES) \citep{parisi2017manifold}. In essence, a parametric policy is used, where each parameter is represented by a Gaussian distribution. Sampling these distributions results in an executable policy that can be applied on our environment. MONES optimises the mean and standard deviation of each parameter such that, whenever we sample from them, we obtain a policy that leads to a different point on the Pareto front. MONES is an iterative process, that repeats three steps:

\begin{enumerate}
    \item Sample a population of policies from our parameters, and execute them on the environment;
    \item Evaluate the quality of these policies using an indicator metric;
    \item Perform a gradient step that improves this indicator using the natural gradient.
\end{enumerate}

Our policy is represented as a small feedforward neural network, where each weight is sampled from its own Gaussian distribution. The network contains a single hidden layer of 50 neurons and, although these weights are correlated with each other, we assume each Gaussian distribution to be independent for the sake of simplicity.

To evaluate the performance of the policies, MONES requires an indicator metric. We closely follow \citep{parisi2017manifold}, where the metric used is a combination of non-dominance ranking and crowding distance. For non-dominance ranking, a rank of 0 is applied for all the non-dominated points of the discovered returns. By removing these solutions from the population, a new set of policies becomes non-dominated. We set the rank for these points to -1. This process is repeated until no points remain to be evaluated. At each iteration the rank is decreased by 1.

This rank is then combined with the crowding distance, which is a metric providing information about the diversity of a frontier:
\begin{itemize}
    \item For all the points of the same rank we compute, for each dimension, the distance between its closest neighbours.
    \item This distance is normalised. Points close to each other will have a crowding distance close to 0, while points at the border of the frontier will have a distance close to 1.
\end{itemize}
Summing these 2 metrics together provides us with an indicator that encourages points to be on the Pareto front, and be as diverse as possible.

The MONES learner is trained for 30 iterations, sampling 50 policies every time. Each policy is executed 10 times. The average return of each policy is evaluated using the non-dominance/crowding-distance metric.

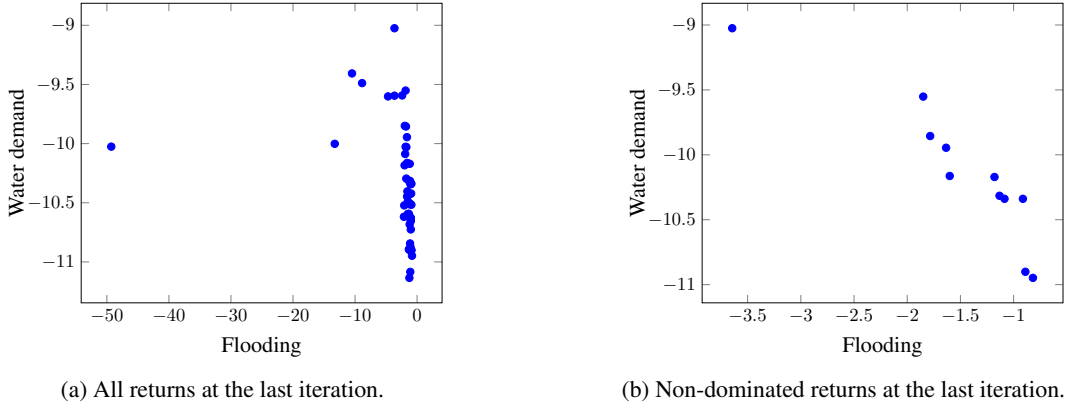
\begin{figure}[h]
    \centering
    \begin{subfigure}{.5\textwidth}
        \centering
        \begin{tikzpicture}[scale=0.7]
        \begin{axis}[
            no markers,
            xlabel={\large Flooding},
            ylabel={\large Water demand}]
        \addplot[only marks, color=blue] table [x=flooding, y=water-demand, col sep=comma] {returns.csv};
        \end{axis}
        \end{tikzpicture}
        \caption{All returns at the last iteration.}
        \label{fig:returns}
    \end{subfigure}%
    \begin{subfigure}{.5\textwidth}
        \centering
        \begin{tikzpicture}[scale=0.7]
        \begin{axis}[
            no markers,
            xlabel={\large Flooding},
            ylabel={\large Water demand}]
        \addplot[only marks, color=blue] table [x=flooding, y=water-demand, col sep=comma] {returns-nd.csv};
        \end{axis}
        \end{tikzpicture}
        \caption{Non-dominated returns at the last iteration.}
        \label{fig:returns-nd}
    \end{subfigure}
    \caption{Comparison of \textit{returns} (left) with \textit{non-dominated returns} (right). In order to enhance presentation, the right plot's horizontal axis was clipped to a smaller interval.}
    \label{fig:mones-returns}
\end{figure}

As can be seen in Figure~\ref{fig:mones-returns}, after 30 iterations of training, the policies sampled from the Gaussian distributions achieve diverse combinations of returns. The right part of the figure shows 11 non-dominated solutions, but the vast majority of the policies (48/50) reach returns reasonably close to the frontier, resulting in a set of diverse, high-quality solutions.

\subsection{Using single-objective subroutines}

Instead of using a dedicated method (MONES) to discover diverse policies, we use an outer loop method. In this particular case, we use Natural Evolution Strategies (NES) as a single objective subroutine. This subroutine is called a number of times, each time with a different utility function, hopefully resulting in different policies that reach different points of the coverage set.

This requires us to know the distribution over user utility functions. We consider a uniform distribution over linear scalarisation functions, i.e. each utility function is a weighted sum, where the weights are uniformly sampled from a 1-simplex (since our problem has two objectives).

Since MONES takes inspiration from NES, both algorithms are quite similar, the main difference being the indicator metric used. While MONES optimises on the combination of ranking and crowding distance, NES optimises on the utility of the return. All other parameters are kept the same as for MONES.

\begin{figure}[h]
    \centering
    \begin{tikzpicture}
        \begin{axis}[
            no markers,
            xlabel={Flooding},
            ylabel={Water demand}]
        \addplot[only marks, color=blue] table [x=flooding, y=water-demand, col sep=comma] {returns-mones.csv};
        \addlegendentry{MONES};
        \addplot[only marks, color=red] table [x=flooding, y=water-demand, col sep=comma] {returns-nes.csv};
        \addlegendentry{NES};
        \end{axis}
    \end{tikzpicture}
    \caption{Comparison of MONES and NES policies.}
    \label{fig:mones-nes}
\end{figure}
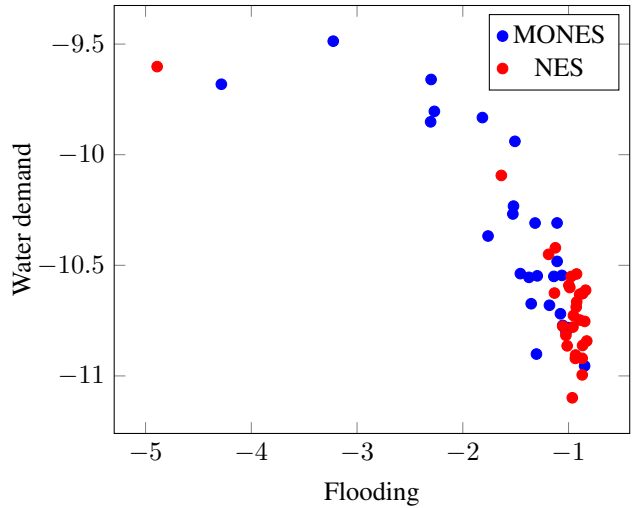

We sampled 30 utility functions, resulting in 30 different NES runs. In Figure~\ref{fig:mones-nes}, we compare the coverage sets found by MONES and NES. In order to have the same number of points for each method, we sampled 30 new policies from our trained MONES and plotted the resulting returns. 

Although the utility functions used by NES were sampled across the whole simplex, it does not result in a spread-out coverage set. The vast majority of returns hover around $(-1.0, -10.7)$, where -1.0 refers to flooding and -10.7 to water demand. In comparison, the policies sampled from MONES result in more spread-out returns. This is because the crowding distance is taken into account in the indicator metric used by MONES, encouraging the returns to be diverse. It is also important to mention that, even though hyperparameters are the same for both methods, NES had to be trained 30 separate times, while MONES only once. Compared to NES, MONES is sample-efficient and results in a more evenly spread-out coverage set. Finally, NES requires us to make assumptions about the distribution over user utility functions, while MONES makes no such assumption.

\subsection{Comparing evaluation metrics}

In order to evaluate the training progression of MONES after each iteration, we use two of the metrics described in Section~\ref{sec:evaluating_performance}. First, we use the hypervolume (Equation~\ref{eq:hypervolume}). This metric requires a reference point, which in this case is set  to the worst return found across the whole training process. Secondly, we use the Expected Utility Metric (EUM, Equation~\ref{eq:eum}). This metric requires a good prior distribution over user utility functions, as well as a good approximation of the solution set. As a prior distribution, we choose the one used for NES, e.g. a uniform distribution over linear scalarisation functions.

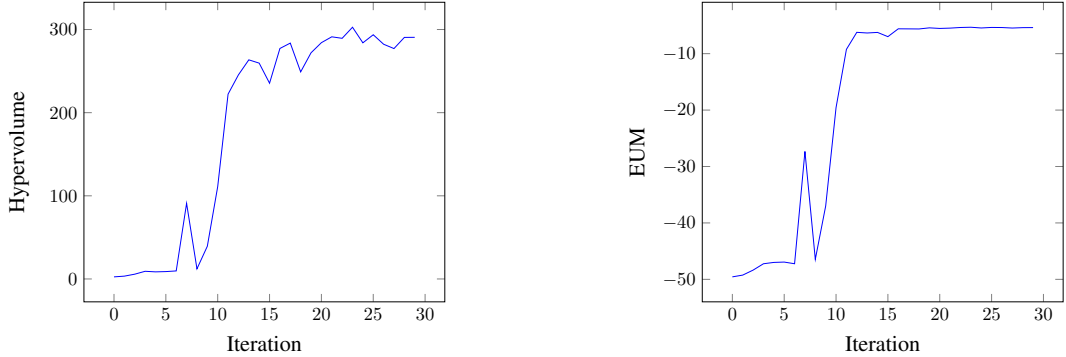
\begin{figure}[h]
    \centering
    \begin{subfigure}{.5\textwidth}
        \begin{tikzpicture}[scale=0.7]
        \begin{axis}[
            no markers,
            xlabel={\large Iteration},
            ylabel={\large Hypervolume}]
        \addplot[color=blue] table [y=hypervolume, col sep=comma] {hypervolume.csv};
        \end{axis}
        \end{tikzpicture}
    \end{subfigure}%
    \begin{subfigure}{.5\textwidth}
        \begin{tikzpicture}[scale=0.7]
        \begin{axis}[
            no markers,
            xlabel={\large Iteration},
            ylabel={\large EUM}]
        \addplot[color=blue] table [y=eum, col sep=comma] {eum-mones.csv};
        \end{axis}
        \end{tikzpicture}
    \end{subfigure}
    \caption{Evaluation of MONES using 2 multi-objective metrics: the Hypervolume (left) and Expected Utility Metric (right).} 
    \label{fig:mones-metrics}
\end{figure}

As we can see in Figure~\ref{fig:mones-metrics}, MONES converges towards our approximation of the optimal solution after 15 iterations, and stays stable for the remainder of the training process. We observe similar trends with the other evaluation metric used (the hypervolume). We also note at iteration 7, that the sudden improvement in EUM is reflected in the hypervolume, although not as drastically. The solution set found in iteration 7 almost doubled the expected user utility, but only resulted in a $30\%$ increase in hypervolume. This shows that, although an increase of hypervolume is correlated with an improvement of the coverage set, it does not reflect the utility of the user.

In conclusion, we tackled the water management problem with a dedicated multi-objective algorithm. By changing the indicator metric of NES to cope with multiple criteria, we discovered a solution set that is more diverse than repeatedly applying its single-objective counterpart, and for which the solutions are more evenly spread out. Moreover, MONES only requires minimal assumptions on the utility function (monotonicity). Finally, the number of required interactions with the environment is vastly improved since, in this case, 30 instances of NES needed to be executed, compared to just a single instance of MONES.

\section{Conclusion, challenges and open questions}
\label{sec:challenges_open}

Recent years have seen significant breakthroughs in the capabilities of sequential decision making agents based on planning or reinforcement learning methods. This has lead to the increasing applications of these agents to complex, real-world problems. However, as illustrated by the motivating examples in Section~\ref{sec:motivating_examples}, these real-world tasks frequently require trade-offs to be made between multiple conflicting objectives. This contrasts with the inherently single-objective nature of the environments such as board  and video games on which the planning and learning algorithms have largely been developed and evaluated. When these single-objective methods are applied to problems which are multi-objective in nature, either some objectives wind up being excluded from consideration, or the objectives are summed together to form a scalar reward. As discussed in Sections \ref{sec:introduction} through \ref{sec:utility_based}, the use of single-objective methods to address multi-objective problems has numerous disadvantages: it forces a priori and uncertain decisions to be made about the desired trade-offs between objectives, it limits the ability to discover multiple policies so as to rapidly adapt to changing preferences, it shifts the responsibility for managing trade-offs from the problem stakeholders to the system developer, and it may result in solutions which fail to maximise user utility. 

While the last decade or so has seen significant achievements in the development of planning and RL algorithms for multi-objective problems (as reflected in Section \ref{sec:survey_algorithms}), it remains a niche area compared to the amount of research on single-objective agents. In addition a number of challenges arise in the context of multiple objectives which do not exist in the single-objective domain. As such there remain a number of areas where additional research and algorithmic development is required. The remainder of this article will present an overview of the topics which we believe to be the most significant and pressing challenges for multi-objective agent research.

\subsection{Lack of multi-objective datasets and benchmarks} \label{sec:conclusion_lack_of_datasets}

Data plays a role in multi-objective decision making (MODM). When solving a given MO problem, data is usually needed to characterise and solve the involved objectives. However, the currently available data may not be sufficient to model some objectives or domains. Whereas this tends not to be an issue for company-oriented research (since companies can usually obtain the required data should it be necessary to achieve its goals), it is often a significant problem for basic research (where the lack of data may make it impossible to study some problems). Some of the challenges faced here include: heterogeneity, availability, and lack of correlation. 

As an example, consider the case of traffic authorities aiming to optimise the control of traffic lights as to minimise competitive objectives like travel time, polluting emissions, and discomfort level. 
To accommodate all these objectives, one needs to deal with all the aforementioned challenges. Data heterogeneity comes into play because the data comes from different sources: data about specific trips come from drivers and passengers; overall traffic statistics come from traffic authorities; fleet demographics come from censoring authorities; CO$_2$ emission profiles come from manufacturers, based on existing fleet; etc. 
Availability refers to the fact that the above information is not openly available, either because of privacy concerns, or due to the lack of data release policies. 
Finally, the data may not be correlated, as is the case of traffic statistics and fleet demographics, which come from different, possibly incompatible sources.

A challenge related to the availability of data is the availability of good benchmark problems for evaluation of MORL algorithms. So far, a limited number of benchmark problems have been proposed for MORL research, and many of those proposed are quite simple (see Section \ref{sec:benchmarks}). Some advantages of the existing benchmark problems are, e.g., that they are simple to understand, experiments can be run in a short time, and optimal solutions to the problems are often known. However, they lack the complexity of many real-world problems that deal with multiple conflicting objectives. Referring to our motivating examples for multi-objective reinforcement learning and planning (see Section~\ref{sec:motivating_examples}), we note that more benchmarks with complex state and action spaces, partial observability, many objectives, multiple agents, and decision making over long time horizons are needed. One approach to quickly increase the number of available MORL benchmarks could be to modify existing single-objective benchmarks, by making their reward functions multi-objective.

In conclusion, these challenges frequently slows down or even hinders research progress on MODM.
Building upon this background, it is of utmost importance for companies and researchers to make their data and benchmarks available. Actions towards this direction include: making your MO problems, data, benchmarks, and baseline implementations available on open platforms; supporting other researchers interested in your problem; negotiating data retention procedures with companies; among others. Without the support of the involved parties, the potential benefits that MODM could bring to our society may not be realised. 

\subsection{Many-objective problems}

Within the field of multi-objective evolutionary optimisation, the task of handling problems with many objectives (usually defined as four or more objectives) has emerged as a distinct sub-field, in recognition that algorithms which work well for a small number of objectives may scale poorly to many objectives \citep{von2014survey,li2015many}. 
So far there has been only minimal work in planning or RL for many-objective problems.
For instance, \citet{zintgraf2018ordered} consider a traffic regulation problem with $11$ objectives (reflecting the delay duration and queue length for different traffic participants and different directions), and how to elicit and model user utility in such a setting using pairwise comparison queries between solutions and Gaussian processes.
\citet{giuliani2014dimensionality} demonstrate a dimensionality-reduction approach in which the original objectives are mapped to a lower dimension using Non-negative Principal Component Analysis, while \citet{yahyaa2014knowledge} examined the performance of bandit algorithms on problems with up to five objectives. However, the development of a broader suite of algorithms for many-objective problems remains an important direction for future work.

\subsection{Multi-agent problems}
\label{sec:conclusion_mas}

Numerous real-world problems involve both multiple actors and objectives that should be considered when making a decision. Multi-objective multi-agent systems represent an ideal setting to study such problems. However, despite its high relevance, it remains an understudied domain, perhaps due to the increasingly complex dimensions involved.

Prior to the recent survey on multi-objective multi-agent decision making \citep{radulescu2020survey}, the literature in this area was rather fragmented and lacked a uniformly accepted framework or set of assumptions to allow for a proper comparison or placement of contributions and to identify gaps in terms of studied settings. Following the taxonomy set out by \citep{radulescu2020survey}, and the links that have been made to suitable solution concepts for MOMADM (briefly discussed in Section~\ref{sec:momas_algos}), we anticipate that the pace of research on multi-objective multi-agent problems will increase in the coming years.

There are countless open challenges brought forward by the MOMADM domain \citep{radulescu2020survey}, ranging from how to develop negotiation strategies for selecting between multiple potential solutions, how equilibria are affected by the choice of the optimisation criteria (SER vs. ESR, Section~\ref{sec:seresr}) and utility functions of the agents, how to learn about the behaviour or objective preferences of other agents, how to deal with sequential or continuous state-action settings. 

Finally, all the observations and remarks regarding the scarce availability of datasets and benchmarks we make in Section~\ref{sec:conclusion_lack_of_datasets} are even more pressing in the case of multi-objective multi-agent settings, rendering the evaluation of MOMADM approaches a challenging task. 

\subsection{Dynamically identifying and adding objectives}

As discussed earlier in Section \ref{sec:design_factors:motivating_scenarios} under the ``review and adjust'' scenario, an analysis of the policy found based on an initial formulation of a problem may reveal the need to modify or extend the objectives considered by the agent in order to find a more acceptable solution. While prior work in the single-objective literature has considered modifying aspects of the problem either during or after learning or planning, such as changes in environmental state dynamics \citep{nagabandi2018learning}, dynamic rewards \citep{devlin2012dynamic}, or introducing new actions \citep{mandel2017add}, obviously the addition of new objectives is unique to multi-objective methods.

Ideally the agent should be able to integrate additional or modified objectives without needing to discard prior learning, and with minimal regret experienced while adjusting its policy. One means by which this might be achieved is to maintain an archive of the agent's experience under its current policy, which can be used to perform offline learning related to updated specifications of objectives without any further interaction with the actual environment. Alternatively, during learning the agent may be able to identify for itself states that could be associated with potential new objectives (for example, states which are highly different in feature space from other states), and create its own rewards associated with these states such that its policy can be rapidly updated should the user define a new objective associated with these states \citep{karimpanal2017identification}.

\subsection{Closing remarks}
The aim of this article is to encourage a wider adoption of multi-objective agent technologies in the development of real-world applications. To this end, we have identified a range of factors which need to be considered when designing a multi-objective solution, as well as reviewing how current multi-objective planning and RL algorithms relate to these factors. In addition, we have provided examples demonstrating how existing methods can be applied to discover suitable agent policies for some simple multi-objective tasks. Our hope is that this article will serve to inspire the future growth of applications based on multi-objective agents.

\vskip 0.2in
\bibliographystyle{abbrvnat}
\bibliography{references}  

\end{document}